\documentclass[conference]{IEEEtran}

\usepackage[hyphens]{url}
\usepackage{graphicx}

\usepackage{xspace}
\usepackage{tabularx}
\usepackage[utf8]{inputenc}
\usepackage{tikz}
\usepackage{booktabs} 
\usepackage{caption}  
\usepackage{subcaption}
\usepackage{enumitem}
\usepackage{listings}
\usepackage{newfloat}   
\usepackage{adjustbox}
\usepackage{amsmath}
\usepackage{amsthm}
\usepackage{multirow}
\usepackage[table]{xcolor}
\usepackage{array}    
\usepackage{algorithm}
\usepackage{algpseudocode}

\newcommand{\stepcircled}[1]{%
  \tikz[baseline=(char.base)]{
    \node[shape=circle,draw=black,fill=black,inner sep=1pt] (char) {\textcolor{white}{\textbf{#1}}};
  }%
}

\newcommand*{\eg}{e.g.,\xspace}
\newcommand*{\ie}{i.e.,\xspace}

\newcommand*{\sys}{\texttt{VirnyFlow}\xspace}
\newcommand*{\interface}{\texttt{VirnyFlow-Interface}\xspace}
\newcommand*{\alpine}{\texttt{Alpine Meadow}\xspace}
\newcommand*{\askl}{\texttt{auto-sklearn}\xspace}
\newcommand*{\flaml}{\texttt{FLAML}\xspace}
\newcommand*{\openbx}{\texttt{OpenBox}\xspace}
\newcommand*{\taskmng}{\texttt{Task Manager}\xspace}
\newcommand*{\wrks}{\texttt{Workers}\xspace}
\newcommand*{\distq}{\texttt{Distributed Queue}\xspace}
\newcommand*{\virny}{\texttt{Virny}\xspace}   

\newcommand*{\lps}{\textit{logical pipelines}\xspace}
\newcommand*{\pps}{\textit{physical pipelines}\xspace}
\newcommand*{\lp}{\textit{logical pipeline}\xspace}
\newcommand*{\pp}{\textit{physical pipeline}\xspace}

\newcommand*{\diabetes}{\texttt{diabetes}\xspace}

\newcommand*{\heart}{\texttt{heart}\xspace}
\newcommand*{\folkemp}{\texttt{folk-emp}\xspace}
\newcommand*{\folkempbig}{\texttt{folk-emp-big}\xspace}
\newcommand*{\folkempall}{\texttt{folk-emp-all}\xspace}
\newcommand*{\folkpc}{\texttt{folk-pubcov}\xspace}

\newcommand{\first}{\emph{(i)}\xspace}
\newcommand{\second}{\emph{(ii)}\xspace}

\newif\ifarxiv
\arxivtrue 
\newcommand{\fullversion}{the full version of the paper~\cite{virnyflow2026extended}}
\newcommand{\fullversionbig}{The full version of the paper~\cite{virnyflow2026extended}}
\newcommand{\inappendix}[1]{\ifarxiv #1\else \fullversion\fi}   
\newcommand{\inappendixbig}[1]{\ifarxiv #1\else \fullversionbig\fi}   
\newcommand{\arxivonly}[1]{\ifarxiv #1\fi}                      

\newtheoremstyle{definition}%
{}{}%
{\itshape}{}%
{\bfseries}{.}%
{ }%
{\thmname{#1}\thmnumber{ #2}\thmnote{ (#3)}}

\definecolor{darkgreen}{rgb}{0.0, 0.5, 0.0} 

\lstdefinelanguage{yaml}{
  keywords={true,false,null,y,n,yes,no},
  keywordstyle=\color{blue},
  comment=[l]{\#},
  commentstyle=\color{gray},
  morestring=[b]',
  morestring=[b]",
  stringstyle=\color{darkgreen},
  moredelim=[l][\color{black}]{-}
}

\DeclareFloatingEnvironment[
  fileext=lol,
  listname={List of Listings},
  name=Listing
]{listing}

\title{VirnyFlow: Optimizing ML Pipelines for Accuracy, Fairness, and Stability at Scale}

\author{
\IEEEauthorblockN{Denys Herasymuk\textsuperscript{1}, Anastasiia Mozghova\textsuperscript{2}, Nazar Protsiv\textsuperscript{3}, Vladyslav Sydorak\textsuperscript{3}, Julia Stoyanovich\textsuperscript{1}}
\IEEEauthorblockA{\textsuperscript{1}New York University, New York, NY, USA \quad
\textsuperscript{2}Taras Shevchenko National University of Kyiv, Kyiv, Ukraine\\
\textsuperscript{3}Ukrainian Catholic University, Lviv, Ukraine\\
\{d.herasymuk, stoyanovich\}@nyu.edu, mozgova\_@knu.ua, \{protsiv.pn, sydorak.pn\}@ucu.edu.ua}
}

\begin{document}

\maketitle
\bstctlcite{BSTcontrol}

\begin{abstract}
Developing machine learning (ML) systems for real-world deployment requires navigating context-dependent trade-offs among accuracy, fairness, stability, and other objectives. Existing AutoML frameworks optimize pipelines efficiently, but they fix the optimization objective up front, leave it outside the developer's control during search, and rarely scale beyond a single node. We present VirnyFlow, a system that optimizes ML pipelines jointly for accuracy, fairness, and stability at scale. A user-defined \emph{evaluation protocol}, with fairness measured over binary and intersectional groups, drives every layer of the optimizer: multi-objective Bayesian optimization of physical pipelines, cost-aware bandit selection of logical pipelines, and multi-criterion pruning. The architecture combines asynchronous execution over Apache Kafka with database-backed experiment management, providing fine-grained parallelism, fault tolerance, and interactive inspection of trade-offs.

On six real-world datasets, VirnyFlow achieves competitive or superior performance compared to state-of-the-art AutoML systems (auto-sklearn, Alpine Meadow, FLAML) under identical resource constraints, scales to 128 workers across four nodes on datasets of up to 2.6M records, and achieves up to 7$\times$ higher speedup than the best-scaling single-node baseline, while maintaining stable accuracy and fairness as parallelism increases. A clinical case study on distribution shift and an IRB-approved user study demonstrate human-in-the-loop navigation of trade-offs in practice: rather than returning a single ``best'' model, VirnyFlow lets data scientists define, inspect, and iteratively refine the objectives of the search to fit their deployment context.
\end{abstract}


\section{Introduction}
\label{sec:intro}

Constructing an ML pipeline for deployment is a multi-stage, multi-objective search problem: null imputation, model selection, and hyperparameter tuning interact, and objectives extend beyond accuracy to fairness across demographic groups, and to the stability of predictions. Which objectives matter, and how they should trade off, depends on the deployment context~\cite{DBLP:journals/cacm/StoyanovichAHJS22,weerts2024can}: a clinical screening tool and a hiring model warrant different metrics and different weights, so these choices belong to the developer, not to the optimizer. Existing AutoML systems largely fix a single objective and return one ``best'' model. Supporting context-specific, multi-objective search instead is an infrastructure problem: the system must evaluate thousands of candidate pipelines efficiently enough to keep a human in the loop as objectives are defined, inspected, and revised.

\lstdefinelanguage{yaml}{
  morekeywords={true,false,null,yes,no},
  sensitive=false,
  morecomment=[l]{\#},
  morestring=[b]',
  morestring=[b]"
}

\lstset{
  basicstyle=\ttfamily\footnotesize,
  language=yaml,
  columns=fullflexible,
  keepspaces=true,
  showstringspaces=false,
  commentstyle=\color{gray},
  keywordstyle=\color{blue},
  stringstyle=\color{teal},
  frame=single,
  breaklines=true,
  escapeinside={(*@}{@*)}
}
\begin{figure}[b!]
\centering
\begin{lstlisting}[language=yaml, caption={Experiment config example: Multi-objective optimization for accuracy, fairness, and stability.}, label={fig:exp_config}]
(*@\textbf{pipeline\_args}@*):
  dataset: "folk_emp"
  sensitive_attrs_for_intervention: ["SEX", "RAC1P"]
  null_imputers: ["median-mode","miss_forest","datawig"]
  fairness_interventions: ["DIR", "AD"]
  models: ["lr_clf","rf_clf","lgbm_clf","gndlf_clf"]
(*@\textbf{optimisation\_args}@*):
  objectives:
   - {metric: "F1", group: "overall", weight: 0.25}
   - {metric: "FNRD", group: "SEX&RAC1P", weight: 0.5}
   - {metric: "Label_Stability", group: "overall",
      weight: 0.25}
  optimizer: {surr_type: "prf", acq_type: "ehvi"}
  max_total_pipelines_num: 100
  num_workers: 32
  num_pp_candidates: 4
  training_set_fractions_for_halting: [0.5, 1.0]
  exploration_factor: 0.5
(*@\textbf{evaluation\_args}@*):
  bootstrap_fraction: 0.8
  n_estimators: 50
  sensitive_attrs: {SEX:'2', RAC1P:['2','3','4','5','6','7','8','9'], SEX&RAC1P: None}
\end{lstlisting}
\end{figure}

As a practical scenario, consider Ann, a data scientist building an employment prediction model that is used to target job training and outreach programs (as in ACSEmployment~\cite{folktables_ding2021}). Ann aims to build an accurate, robust, and fair (\eg in the sense of error rate parity) ML pipeline, which requires making a series of context-dependent design choices.

First, she must decide which performance dimensions are appropriate in this context, such as fairness across sex, race, and their intersections, and stability, and encode them into the optimization objective; optimizing one metric rarely improves others~\cite{shades, whang2023data, karl2023multi} because of non-convex fairness constraints and tuning issues~\cite{whang2023data}. Second, objectives must span the full ML lifecycle, reflecting how errors introduced during data collection propagate through preprocessing (\eg imputation), model selection, and tuning~\cite{DBLP:journals/cacm/StoyanovichAHJS22, shades, whang2023data, suresh2021framework}. Third, exploring \emph{tens of thousands} of pipeline variants demands distributed resources, early pruning, and careful experiment management, especially when objectives are revised mid-course.

Motivated by these limitations in existing systems, we introduce \sys to assist Ann in navigating this design space, framing development as \emph{exploring and comparing alternative pipeline configurations under explicit, revisable objectives rather than optimizing a fixed criterion}. Concretely, an alternative in this space is a full pipeline configuration: Ann may compare a pipeline that composes \texttt{missForest} imputation, the Disparate Impact Remover intervention, and a gradient-boosted model against one that composes median-mode imputation, Adversarial Debiasing, and logistic regression. The two alternatives differ in F1, in false negative rate difference across sex, race, and their intersection, and in training cost; neither dominates the other. We preview the system's flexibility in Listing~\ref{fig:exp_config}, which shows an experiment configuration Ann may specify for her scenario.

\subsection{Background and Related Work}
\label{sec:intro:related}

ML pipeline optimization has produced a diverse ecosystem of \textit{AutoML} tools~\cite{baratchi2024automated}, from hyperparameter optimization and neural architecture search to broad-spectrum approaches to the combined algorithm selection and hyperparameter optimization problem~\cite{thornton2013auto} via Bayesian optimization, bandits, and evolutionary search~\cite{feurer2022auto,swearingen2017atm,shang2019democratizing}.

Incorporating fairness, stability, and other criteria into pipeline optimization requires spanning the entire ML lifecycle~\cite{DBLP:journals/cacm/StoyanovichAHJS22,whang2023data,suresh2021framework}: errors and biases introduced upstream, \eg through imbalanced sampling during data collection~\cite{feldman2015certifying,zhang2018mitigating}, propagate through imputation, model selection, and tuning, so pipeline stages cannot be optimized in isolation. This motivates \emph{multi-stage}, \emph{multi-objective} optimization.

Recent AutoML efforts incorporate fairness into the optimization loop~\cite{komala2024fair,robertson2024human,wu2021fairautoml,wang2021flaml}, but with three capability gaps: they optimize model tuning and selection rather than the full pipeline; they do not support stability or uncertainty as objectives~\cite{feder2023variance}; and they operate over binary, non-intersectional groups with a narrow set of fairness metrics~\cite{barocas2023fairness}. They also fix objectives up front, with no mechanism for revising them during development. Adjacent work on human-in-the-loop pipeline design~\cite{redyuk2024assisted,neutatz2024automl}, scalable ML systems~\cite{george2020scalable}, and LLM-based agents~\cite{hong2025data} similarly lacks configurable evaluation protocols, intersectional fairness, and context-sensitive optimizer selection~\cite{weerts2024can}.

\textbf{In summary,} no existing system combines context-specific, multi-objective optimization across the pipeline lifecycle with the iterative, developer-driven refinement that pipeline construction requires in practice~\cite{weerts2024can}. \alpine~\cite{shang2019democratizing} comes closest through its emphasis on interactivity, and \sys builds on two of its ideas. First, we combine multi-armed bandits with Bayesian optimization to improve exploration and interactivity, integrating an \textit{evaluation protocol} that supports flexible objective specification, multi-dimensional measurement, and multi-objective optimization. Second, we extend \alpine{}'s query optimization, pruning, and \textit{adaptive pipeline selection} (Algorithm~\ref{alg:adaptive_pipeline_selection}) to the multi-objective setting.

\subsection{Scope and Contributions}
\label{sec:intro:scope}

We present \sys\footnote{\url{https://github.com/denysgerasymuk799/virny-flow}}, a system that optimizes ML pipelines jointly for accuracy, fairness, and stability at scale. To our knowledge, \sys is the first system in which a user-defined \textit{evaluation protocol} drives every layer of a distributed multi-stage pipeline optimizer. The protocol specifies the metrics, groups, and weights that matter for a task. It parameterizes the multi-objective Bayesian optimization (MOBO) that tunes physical pipelines, the cost-aware bandit that selects logical pipelines, and the multi-criterion pruning rule. We call the resulting environment a \textit{design space}: rather than returning a single ``best'' pipeline, \sys supports systematic exploration of alternative pipeline configurations, evaluation protocols, and optimization strategies.

We focus on supervised ML pipelines over tabular data, the dominant setting in domains such as hiring and healthcare. The computational challenge lies in the volume of pipeline evaluations rather than in the size of any single dataset: thousands of candidate pipelines per task, each potentially requiring bootstrap-based stability estimation over dozens of trained models, must be evaluated at interactive speed. Our experiments address three research questions:

\begin{enumerate}[label=\textbf{RQ\arabic*},leftmargin=3.2em]
    \item Can \sys optimize ML pipelines according to different multi-objective criteria (Section~\ref{sec:exp1})?

    \item How does \sys compare to state-of-the-art AutoML systems in terms of performance ({\textbf{RQ2.1}}) and scalability ({\textbf{RQ2.2}}) (Section~\ref{sec:comparison})?

    \item What is the sensitivity of \sys to different configuration settings (Section~\ref{sec:exp4})?
\end{enumerate}

Our contributions are:

\textbf{\stepcircled{1}: Protocol-driven, multi-stage, multi-objective optimization (Sections~\ref{sec:overview} and~\ref{sec:optimization}).} The evaluation protocol is a first-class object: it is specified in the experiment configuration and applied consistently in optimization, pipeline selection, and pruning. This design extends \alpine{}'s cost model and adaptive pipeline selection to the multi-objective setting, and tunes preprocessing, fairness interventions, and models jointly rather than per stage.

\textbf{\stepcircled{2}: A scalable distributed architecture (Section~\ref{sec:execution}).} \sys combines asynchronous task execution over Apache Kafka, database-backed experiment management with fault tolerance, and fine-grained parallelism, alongside an interactive visualization interface. The same infrastructure that provides scale also lets developers inspect optimization dynamics and revisit objectives as they evolve.

\textbf{\stepcircled{3}: Empirical evaluation (Sections~\ref{sec:experiments} and~\ref{sec:user_study}).} On six real-world benchmarks with up to 2.6M records, \sys achieves competitive or superior performance compared to state-of-the-art AutoML systems under identical resource constraints, and scales to 128 workers across four nodes with up to 7$\times$ higher speedup than the best-scaling baseline. A clinical case study on distribution shift and an IRB-approved user study show that data scientists can reason about context-specific trade-offs using the system.
\section{System Design and Model Development Workflow}
\label{sec:system_design}

This section presents the system design and model development workflow of \sys, and explains how its architecture supports context-sensitive, human-in-the-loop model development by making key design choices explicit, inspectable, and revisable throughout ML pipeline optimization. The design space supports flexible performance criteria, multi-stage and multi-objective optimization, scalable execution, and comprehensive experiment management. 

\subsection{Human-in-the-Loop Design Space}
\label{sec:overview}

We define a \emph{design space} as the set of alternative pipeline configurations, evaluation protocols, and optimization strategies that a data scientist can explore, compare, and refine under task-specific objectives and constraints.  As discussed in Section~\ref{sec:intro}, this design process encompasses multiple challenges. The user workflow below illustrates how \sys's features assist a data scientist throughout this process.

A central design decision in this workflow is the specification of an \emph{evaluation protocol} that defines which performance dimensions matter for a given problem and how they are measured, including which populations are evaluated and how trade-offs among objectives are expressed. Unlike many AutoML systems, which treat evaluation criteria as fixed or implicit, \sys makes protocol definition an explicit configuration step, specified by the developer and revisable between runs.

\textbf{Step~\stepcircled{1}: Define evaluation protocol.} The data scientist analyzes the problem context and selects optimization metrics, such as accuracy, fairness, and stability, together with the populations over which they are measured and the weights that express their relative priority. To support this step, \sys builds on \virny\arxivonly{~\cite{herasymuk2024responsible_real}}, an open-source library for profiling model performance across accuracy, stability, uncertainty, and fairness dimensions. Protocols are specified in the experiment configuration, making metrics, subgroups, and weights explicit and revisable (see Listing~\ref{fig:exp_config}).

\textbf{Step~\stepcircled{2}: Configure experiment.} The data scientist selects and configures relevant components for each stage of the ML pipeline from \sys’s \textit{search space}
, which includes various null imputers, fairness interventions, and ML models. They may also provide custom components. These selections, along with the \textit{evaluation protocol} and execution budget, encode the data scientist’s design intent and are specified in an experiment configuration file. 

\textbf{Step~\stepcircled{3}: Run experiment.} The data scientist deploys \sys across multiple compute nodes using the provided deployment scripts. During execution, the system distributes and evaluates candidate pipelines efficiently
, while the interactive visualization interface allows to monitor progress, interpret optimization dynamics and identify unexpected or undesirable behaviors early, explore trade-offs between objectives, and debug any emerging issues. 

\textbf{Step~\stepcircled{4}: Refine experiment.} If the resulting pipelines do not adequately reflect the desired trade-offs, the data scientist can improve it by experimenting with different optimizers (see \inappendix{Appendix~\ref{apdx:search_space}}), adjusting the feature set, or modifying pipeline components. The \interface supports side-by-side comparison of experiments, as all results are persistently stored in a database. Once a pipeline is deemed satisfactory given the specified objectives and constraints, the user can extract its configuration and deploy it for real-world inference.

\textbf{Step~\stepcircled{5}: Continue experimentation.} As new data, features, or algorithms become available, the data scientist—or a new team member—can continue refining the pipeline using \sys as objectives, constraints, or assumptions evolve. With access to detailed experiment logs and built-in comparison tools, they can effectively build on earlier work, revisit prior decisions, and track improvements over time.

\textbf{Division of labor between human and system.} Human intervention in \sys is opt-in, confined to  decision points that require judgment rather than distributed throughout the search. Given a specified evaluation protocol, optimization runs fully autonomously: logical pipeline selection, hyperparameter tuning, and pruning require no human input. Judgment enters when the protocol is defined (Step~\stepcircled{1}), at inspection checkpoints (Steps~\stepcircled{3}--\stepcircled{4}), and when objectives are revised (Step~\stepcircled{5}). We also note the scope of the transparency claim: \sys makes the \emph{optimization process} transparent and steerable through explicit protocols, Pareto frontiers, and experiment histories; it makes no claim about the interpretability of the learned models themselves, an orthogonal concern addressed by the XAI literature.

\subsection{Navigating the Design Space: Optimization and Trade-offs}
\label{sec:optimization}

The goal of \sys’s optimization machinery is not to converge on a single optimal pipeline, but to efficiently \emph{navigate} the space of alternative pipeline configurations under multiple, potentially conflicting objectives. To support context-sensitive and human-in-the-loop model development, \sys surfaces trade-offs among accuracy, fairness, stability, and other dimensions, while remaining interactive and scalable to support iterative exploration.

\begin{figure}[t!]
    \centering
    \includegraphics[width=\linewidth]{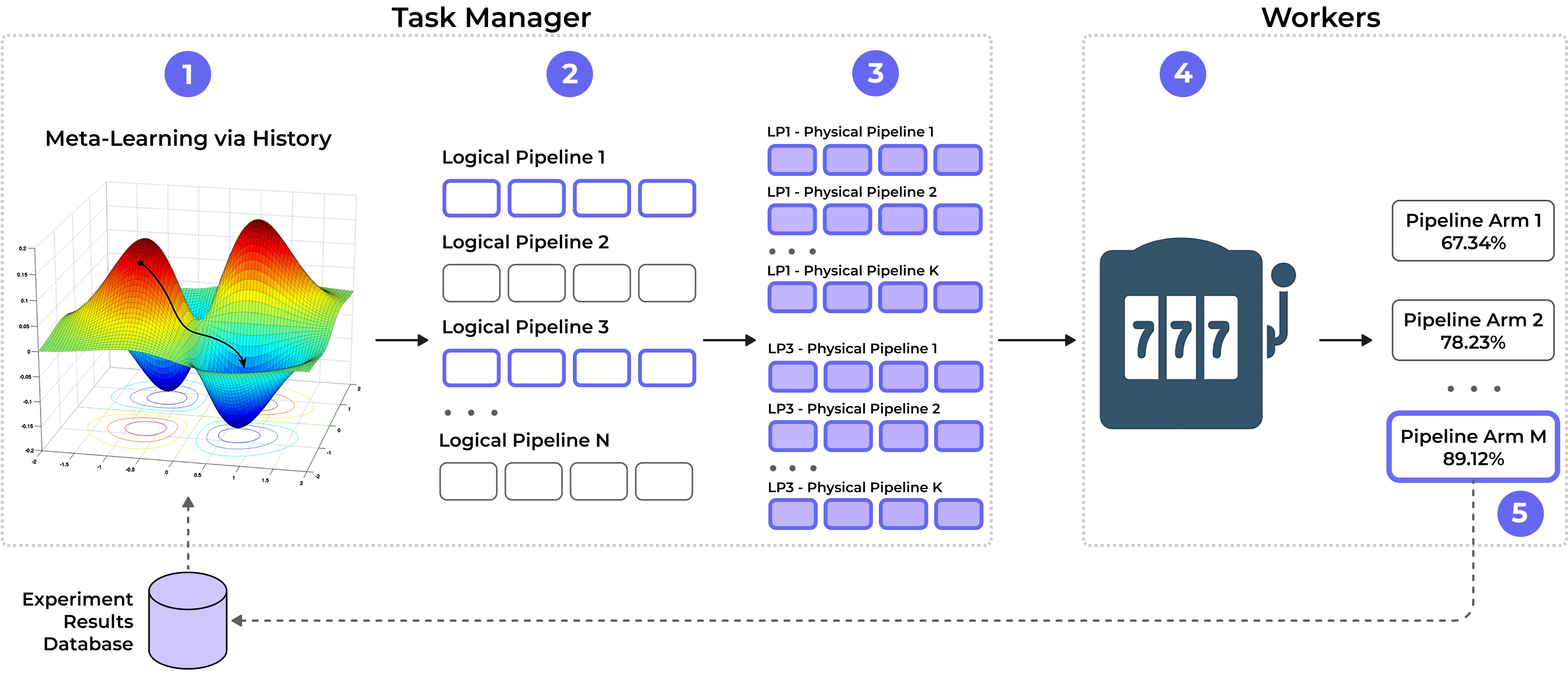}
    \caption{Optimization process: (1) search space definition, (2) logical pipeline selection, (3) physical pipeline selection, (4) pipeline evaluation \& pruning, (5) search space model update.}
    \label{fig:workflow}
\end{figure}

\looseness=-1
Figure~\ref{fig:workflow} illustrates the high-level optimization loop in \sys. Given an experiment configuration (\eg Listing~\ref{fig:exp_config}), the system proceeds through an iterative process: (1) it constructs a search space of candidate \textit{logical pipelines}, which represent high-level compositions of pipeline stages such as data preprocessing, fairness interventions, and model evaluation, each with multiple variants and associated hyper-parameter domains; (2) it prioritizes which logical pipelines to explore next using a \textit{cost model} informed by previously observed execution results; (3) selected logical pipelines are instantiated into multiple \textit{physical pipelines}, which correspond to concrete instantiations with fixed hyperparameters, via Bayesian optimization, with additional random candidates introduced to mitigate local optima; (4) worker nodes execute and prune these physical pipelines using adaptive pipeline selection with multi-criterion optimization to allocate resources efficiently and reduce training time; and (5) execution outcomes are stored and used to continuously refine the cost model for selecting the next promising logical pipeline and update the BO model for physical pipeline selection. This loop repeats continuously, allowing users to inspect intermediate results, revise objectives or configurations, and steer subsequent exploration.


\textbf{Multi-stage, multi-objective optimization.}  To tune physical pipelines, \sys employs multi-objective Bayesian optimization (MOBO)~\cite{garnett2023bayesian}, jointly optimizing multiple objectives across pipeline stages. Unlike approaches that tune stages independently or collapse objectives into a single aggregate, \sys tunes preprocessing, modeling, and fairness interventions together under a user-specified evaluation protocol. Objective definitions and weights are taken directly from this protocol (Section~\ref{sec:overview}), so optimization remains aligned with the problem context.


To explore this complex search space without prematurely converging to local optima, each selected \textit{logical pipeline} is instantiated into multiple \textit{physical pipelines}, representing concrete configurations with fixed hyperparameters. These are generated using Bayesian optimization (BO), with additional random candidates introduced to promote exploration. Through its integration with \openbx~\cite{li2021openbox}, \sys supports multiple BO optimizers and acquisition strategies—including EI~\cite{movckus1975bayesian}, EHVI~\cite{emmerich2006single}, and MESMO~\cite{belakaria2019max}—allowing users to select optimization behavior appropriate for a given evaluation protocol. Additional details on the BO formulation are provided in \inappendix{Appendix~\ref{sec:bo}}.


\textbf{Prioritizing exploration under budget constraints.} Exploring thousands of candidate pipelines requires careful prioritization. \sys uses a bandit-based selection strategy to choose promising logical pipelines for further exploration based on a scoring model that integrates observed performance, uncertainty, and execution cost. Logical pipeline selection is formulated as a multi-armed bandit process that iteratively (1) selects a logical pipeline for execution by sampling proportionally to its score, (2) stores execution outcomes and costs in the experiment history, and (3) updates pipeline scores before repeating the process. The \textit{exploration factor} controls the probability of selecting the currently top-scoring \lp or choosing an unfinished \lp for execution, see \inappendix{Algorithm~\ref{alg:next_logical_plan} in Appendix~\ref{apdx:execution}}.
The score of a logical pipeline is defined as:

\begin{equation}
s = \sum_{i=1}^{n} w_i \cdot \mu_i + \frac{\theta}{c} \cdot \sum_{i=1}^{n} w_i \cdot \delta_i
\label{eq:score}
\end{equation}

\noindent where $n$ is the number of optimization objectives, $w_i$ denotes user-defined weights, $\mu_i$ and $\delta_i$ are the mean and standard deviation of pipeline quality across objectives, $c$ is the estimated execution cost, and $\theta$ controls the degree of exploration by favoring pipelines with higher uncertainty. Weights encode user-specified preferences and are not optimized; scalarization supports efficient search and pruning, while Pareto frontiers are exposed to users for explicit inspection of trade-offs. This formulation balances strong-performing pipelines with uncertain alternatives while accounting for execution cost. Importantly, prioritization is driven entirely by task-specific observations collected during the current optimization run and does not rely on cross-dataset transfer, keeping optimization behavior task-scoped.

\textbf{Incremental evaluation and pruning.} To maintain interactivity and reduce wasted computation, \sys incorporates adaptive pruning during pipeline evaluation. Candidate pipelines are evaluated incrementally on increasing fractions of the training data, and poorly performing pipelines are terminated early when they are unlikely to satisfy the evaluation protocol. Algorithm~\ref{alg:adaptive_pipeline_selection}, \textit{Adaptive Pipeline Selection} (APS), implements this bandit-based pruning strategy; it is based on \alpine~\cite{shang2019democratizing}, which we extend to support multiple optimization objectives. In lines 1--2, the algorithm splits the dataset $\mathcal{D}$ into training and test sets, then subsamples the training set in increments (\eg starting at 50\% and increasing by 10\%), reducing execution costs similarly to successive halving~\cite{jamieson2016non}. The halting criterion terminates a pipeline whose partial training error exceeds the best test error observed so far. We extend this principle to the multi-objective setting by computing $err_{train}$ and $err_{test}$ as weighted sums of errors across objectives, using the user-defined weights from the evaluation protocol (\eg Listing~\ref{fig:exp_config}); this enables efficient elimination of unpromising alternatives without collapsing the search to a single objective. The variable $observation_{test}$ stores non-aggregated performance metrics of a pipeline, which are later used to update the score of the \lp.

\begin{algorithm}[tb]
\caption{Adaptive Pipeline Selection (APS)}
\label{alg:adaptive_pipeline_selection}
\begin{algorithmic}[1]
\Require Pipeline $pipeline$, dataset $\mathcal{D}$
\Ensure Score (negation of error), test objective metrics
\State Split $\mathcal{D}$ into $\mathcal{D}_{train}$ and $\mathcal{D}_{test}$
\State Split $\mathcal{D}_{train}$ into equal-sized $\mathcal{D}_{train}^{1}, \dots, \mathcal{D}_{train}^{N}$
\For{$i \gets 1$ \textbf{to} $N$}
    \State Train $pipeline$ on $\mathcal{D}_{train}^{1 \dots i}$
    \State $err_{test},\ observation_{test} \gets$ Test $pipeline$ on $\mathcal{D}_{test}$
    \If{$err_{test} < err_{best}$}
        \State $err_{best} \gets err_{test}$
    \EndIf
    \State \textbf{yield} $err_{test},\ observation_{test}$
    \State $err_{train},\ observation_{train} \gets$ Test $pipeline$ on $\mathcal{D}_{train}^{1 \dots i}$
    \If{$err_{train} > err_{best}$}
        \State \Return $err_{test},\ observation_{test}$
    \EndIf
\EndFor
\State \Return $err_{test},\ observation_{test}$
\end{algorithmic}
\end{algorithm} 

\textbf{Interactivity.} All optimization components are embedded in the distributed execution framework described in Section~\ref{sec:execution}, which supports fine-grained parallelism and asynchronous execution. This infrastructure allows \sys to evaluate many pipelines concurrently while presenting promising and intermediate results to users as soon as they become available, supported by interactivity-aware scoring and adaptive pruning logic. A dedicated visualization interface\footnote{The visualization interface is available on the HuggingFace platform and can be accessed at: \url{https://huggingface.co/spaces/virnyflow/user-study}.} exposes optimization dynamics, trade-offs, and comparative performance across pipelines, enabling users to monitor execution progress, explore performance across multiple dimensions, analyze trade-offs, and compare pipelines, optimizers, and system configurations over time. Together, these capabilities allow users to interpret results, detect undesirable behavior, and decide when and how to revise the experiment configuration; detailed descriptions and illustrations of the interface are provided in \inappendix{Appendix~\ref{apdx:viz_interfaces}}.

\textbf{In summary,} by combining multi-stage, multi-objective optimization with adaptive prioritization, pruning, and distributed execution, \sys makes large design spaces practically navigable by human users, and is explicitly structured to support exploration, comparison, and revision of alternatives rather than black-box optimization.

\subsection{Scalable Distributed Execution}
\label{sec:execution}

\begin{figure}[t!]
  \centering
  \includegraphics[width=\linewidth]{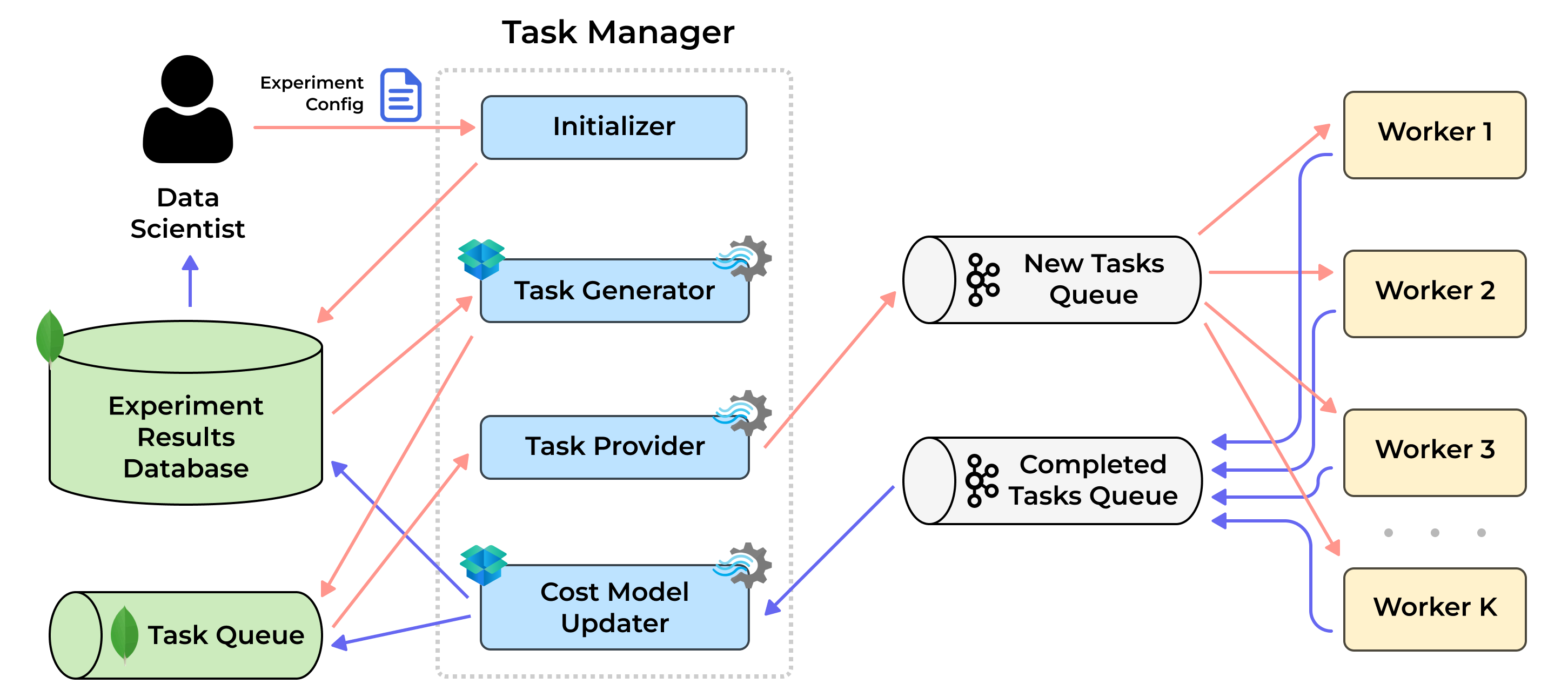}
  \caption{The \sys architecture. MongoDB components shown in green, Kafka queues in gray, workers in yellow, and \texttt{Task Manager} modules in blue. \texttt{Task Provider}, \texttt{Task Generator}, and \texttt{Cost Model Updater} use OpenBox for multi-stage, multi-objective optimization.}
  \label{fig:architecture}
\end{figure}

To enhance scalability and efficiency, \sys combines distributed execution, fine-grained parallelism, asynchronous communication, and asynchronous programming. Figure~\ref{fig:architecture} presents the architecture of \sys, which is implemented in Python and consists of \taskmng, \wrks, \distq, and an external database (MongoDB~\cite{mongodb}). Within \taskmng, there are four key components: \texttt{Initializer}, \texttt{Task Generator}, \texttt{Task Provider}, and \texttt{Cost Model Updater}, with the last three functioning as asynchronous data processors.

The execution process begins when a user provides an \textit{experiment config} and a search space for tuning to \taskmng. The \texttt{Initializer} then instantiates all \lps in the external database. Next, the \texttt{Task Generator} initializes an individual \textit{BO-advisor} for each \lp, applies the cost model to select a promising \lp (Section~\ref{sec:optimization}), and uses the \textit{BO-advisor} to generate an optimal set of hyper-parameters, forming a \pp. This \pp is then packaged into a \textit{task} and stored in the \texttt{Task Queue} in the database.

Tasks are initially staged in the external database rather than sent directly to the \distq for \emph{fault tolerance}: if a system failure occurs and any component shuts down, execution progress and results remain intact in the database, and a user can restart \sys and resume execution from the last saved state. The \texttt{Task Provider} moves tasks from the \texttt{Task Queue} to the \distq.

To enable efficient communication between \taskmng and \wrks, \sys uses \distq, a distributed event queue built on Apache Kafka~\cite{kafka} with two topics: the \texttt{New Tasks Queue}, containing tasks for workers to execute, and the \texttt{Completed Tasks Queue}, storing execution results (\textit{observations}). Fine-grained task parallelism combined with asynchronous messaging ensures effective resource use: the cost of executing a single \textit{task} is relatively low, and each \texttt{Worker} retrieves the next available task from the queue as soon as it completes the previous one. Execution results are processed by the \texttt{Cost Model Updater}, which reads from the \texttt{Completed Tasks Queue} and updates both the global cost model and the corresponding \textit{BO-advisor} for the associated \lp; all execution progress and results are persisted in the database. \taskmng components are implemented with asynchronous programming (\texttt{asyncio}) for non-blocking execution, and the \texttt{Cost Model Updater} uses just-in-time compilation via \texttt{numba} to rapidly recompute \lp scores as new \textit{observations} arrive. Additional implementation details, including task-queue sizing and database optimizations, are in \inappendix{Appendix~\ref{apdx:execution}}.


\section{Experiments}
\label{sec:experiments}

Our experiments address the three research questions posed in Section~\ref{sec:intro:scope}: multi-objective optimization under context-specific criteria (\textbf{RQ1}, Section~\ref{sec:exp1}), comparison to state-of-the-art AutoML systems in performance and scalability (\textbf{RQ2}, Section~\ref{sec:comparison}), and sensitivity to configuration settings (\textbf{RQ3}, Section~\ref{sec:exp4}).

\subsection{Experimental Setup}
\label{sec:exp_setup}

\begin{table*}[t!]
    \centering
   \footnotesize
    \caption{Datasets, tasks, and deployment context.}
    \label{tab:dataset-info}
    \setlength{\tabcolsep}{4pt}
    \begin{tabular}{l l r r c}
        \toprule
        \textbf{name} & \textbf{domain / task / context} & \textbf{\# tuples} & \textbf{\# attrs} & \textbf{sensitive attrs} \\
        \midrule
        \diabetes & healthcare: diabetes prediction for screening & 952 & 17 & sex \\
        \folkemp & hiring: employment prediction with fairness considerations & 15,000 & 16 & sex, race \\
        \folkpc & public coverage: insurance eligibility prediction & 50,000 & 19 & sex, race \\
        \heart & healthcare: cardiovascular risk prediction for screening & 70,000 & 11 & sex \\
        \folkempbig & hiring: employment prediction, single-state subsample & 200,000 & 16 & sex, race \\
        \folkempall & hiring: employment prediction, all US states & 2,590,315 & 16 & sex, race \\
        \bottomrule
    \end{tabular}
\end{table*}

\textbf{Datasets and tasks.} We conduct experiments on six datasets spanning hiring, healthcare, and public insurance, summarized in Table~\ref{tab:dataset-info} (see \inappendix{Appendix~\ref{apdx:datasets}} for details). Each dataset is paired with a deployment scenario that motivates the choice of evaluation objectives, rather than treating metrics as intrinsic dataset properties. Datasets smaller than 1,000 rows are split 70\%/30\% into training and test sets; otherwise, 80\%/20\%.

\textbf{Baselines.} We compare against three state-of-the-art, broad-spectrum AutoML systems highlighted in recent benchmarks and surveys~\cite{baratchi2024automated,neutatz2024automl}: \alpine~\cite{shang2019democratizing}, \askl~\cite{feurer2022auto}, and \flaml~\cite{wang2021flaml}. Among the recent fairness-aware, human-in-the-loop, and LLM-based systems of Section~\ref{sec:intro:related}, we selected \flaml because the others lack public code, optimization-time fairness or stability support, intersectional group handling, or a scalable architecture. For a fair comparison, we use the system configurations as specified in their original papers and standardize the search space across all systems with a common set of ML models and hyperparameters (see \inappendix{Appendices~\ref{apdx:model_types} and \ref{apdx:baselines}}).

\textbf{Optimization metrics.} For each dataset, we select metrics based on its deployment scenario (Table~\ref{tab:dataset-info}); these define the \textit{evaluation protocol} for \sys, which uses EHVI~\cite{emmerich2006single} to tune them jointly. \flaml also supports multi-objective optimization, while the other baselines optimize only for F1. We report \emph{F1} (more reliable than accuracy under class imbalance); average \emph{Label Stability (LS)}~\cite{darling2018toward, arifkhan2023stability} over $B=50$ bootstrapped models; and the error disparity metrics \emph{TPRD}, \emph{TNRD}, \emph{FNRD}, and \emph{SRD}. Efficiency is measured by \textit{runtime} (total time to evaluate a fixed number of pipelines) and \textit{speedup} relative to \sys{}'s single-worker runtime, which is also the baseline for the other systems' speedup (on \folkempall, relative to \sys{}'s 4-worker runtime, the smallest configuration feasible at that scale). Metric definitions and infrastructure details are provided in \inappendix{Appendices~\ref{apdx:exp:metrics} and~\ref{apdx:infrastructure}}.

\subsection{Functionality of \sys}
\label{sec:exp1}


\begin{figure}[b!]
    \centering
    \begin{subfigure}{\linewidth}
        \centering
        \includegraphics[width=\linewidth]{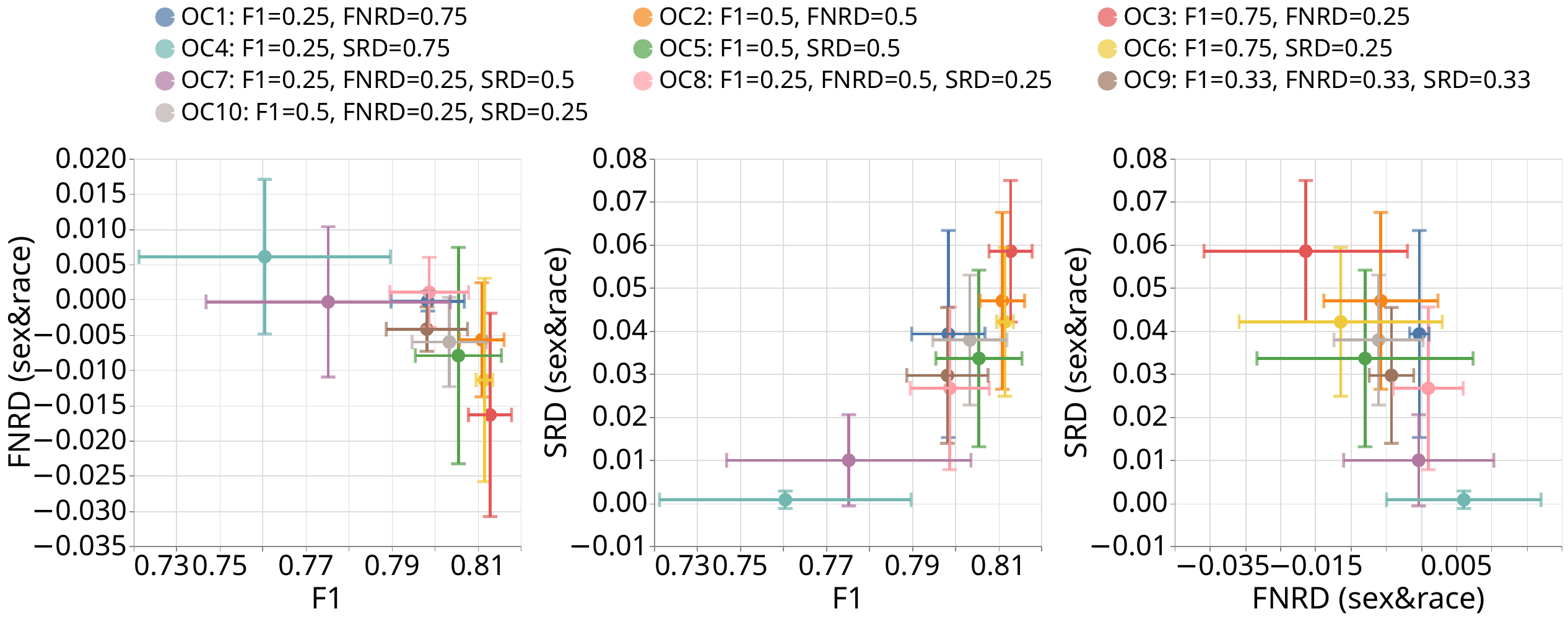}
        \caption{\folkemp}
        \label{fig:exp1_folk_emp}
    \end{subfigure}
    \vspace{0.2cm}
    \begin{subfigure}{\linewidth}
        \centering
        \includegraphics[width=\linewidth]{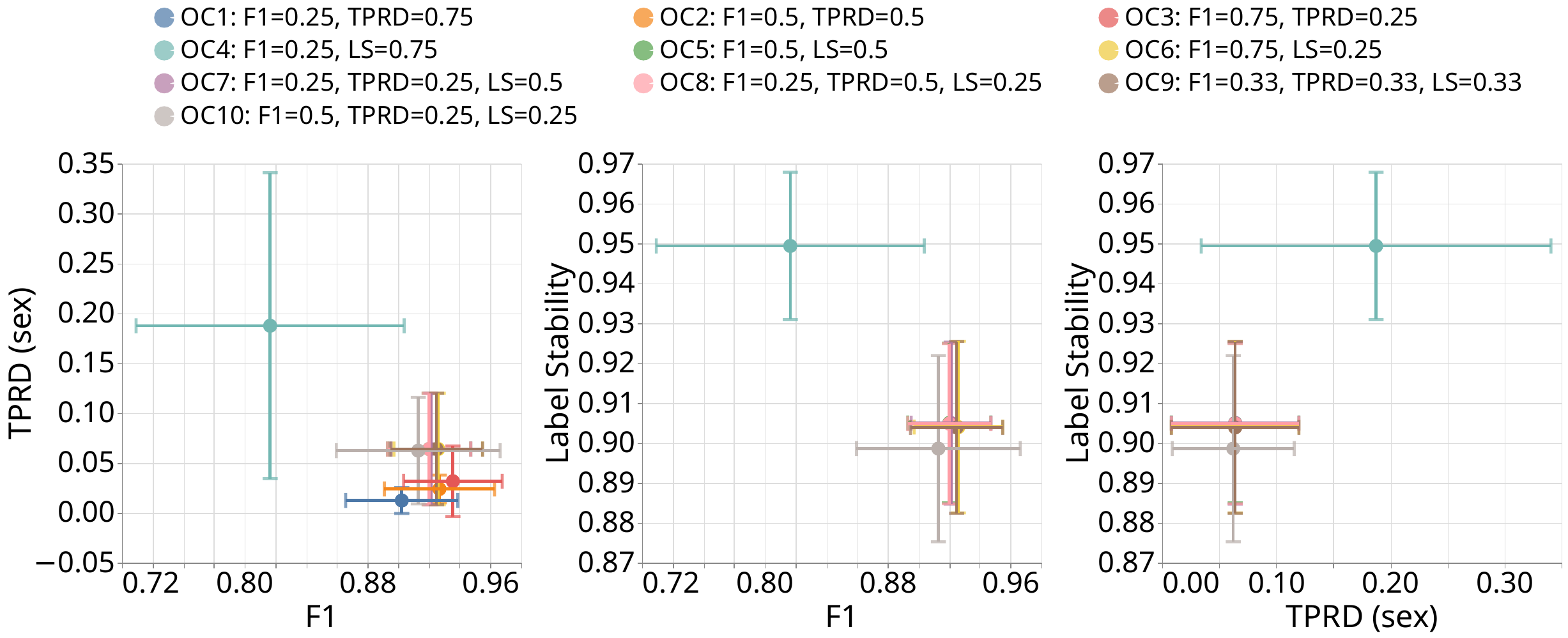}
        \caption{\diabetes}
        \label{fig:exp1_diabetes}
    \end{subfigure}
    \caption{Performance of multi-objective tuning on (a) \folkemp and (b) \diabetes. Points are optimization criteria (OC). Values closer to 0 indicate better fairness.}
    \label{fig:exp1_main}
\end{figure}

In this section, we evaluate the functionality of \sys, demonstrating its ability to optimize ML pipelines based on multiple objectives (\textbf{RQ1}). We conduct case studies on three datasets of varying sizes and domains: \diabetes, \folkemp, and \folkpc. The search space includes four models (\texttt{lr\_clf}, \texttt{rf\_clf}, \texttt{lgbm\_clf}, and \texttt{gandalf\_clf}), along with their respective hyperparameter grids (details in \inappendix{Appendix~\ref{apdx:model_types}}), and does not include feature engineering and fairness-enhancing interventions in this evaluation.

To highlight \sys{}’s capability for multi-objective optimization, we define task-specific objective sets grounded in the deployment contexts of Table~\ref{tab:dataset-info}. To assess how \sys handles prioritization among objectives, we evaluate multiple weight configurations (summing to 1) for each task. Each task is evaluated using 10 optimization criteria. Optimization runs for up to 800 trials (\ie candidate pipelines) per task and seed, and the selected pipeline is used for evaluation. The number of suggestions per \lp is set to $k=2$; the fractions for halting are \{0.7, 1.0\}.

Figures~\ref{fig:exp1_folk_emp} and~\ref{fig:exp1_diabetes} show the mean / std. of the best-performing ML pipelines optimized using different criteria on \folkemp and \diabetes, respectively, over ten random seeds; corresponding results for \folkpc appear in \inappendix{Figure~\ref{fig:exp1_folk_pubcov} in Appendix~\ref{apdx:exp_additional}}. TPRD, FNRD, and SRD values closer to zero indicate better fairness. Each metric is reported regardless of whether it was included in the optimization objective, allowing comparison across settings.

Figure~\ref{fig:exp1_folk_emp} shows that \sys effectively optimizes both fairness and accuracy, even for intersectional groups (\eg sex \& race). In Figure~\ref{fig:exp1_folk_emp}-left, optimization criterion OC1 (favoring FNRD over F1) achieves perfect fairness (FNRD = 0.0) with only a small accuracy drop (F1 decreases by 0.012). Similarly, Figure~\ref{fig:exp1_folk_emp}-center shows that OC4 (prioritizing SRD over F1) achieves SRD = 0.0 but at a greater F1 cost (drop of 0.05), suggesting SRD is harder to optimize on \folkemp. Importantly, \sys can jointly optimize all three metrics: OC9 (equal weights) yields F1 = 0.80, FNRD = -0.004, and SRD = 0.03, each close to the respective best, and demonstrates strong trade-off performance suitable for production use.

On \folkpc, where SRD is optimized separately for sex and race alongside F1, \sys optimizes all three metrics but exposes a clear fairness--accuracy trade-off: jointly optimizing SRD for both groups (OC7) incurs a notable F1 drop ($\approx$0.11), while equal weighting (OC9) loses only $\approx$0.02 F1 but yields significantly worse SRD for race (by 0.06); which to prefer depends on the problem's context and constraints. On \diabetes, where F1, LS, and TPRD for sex are optimized under different weightings, \sys effectively tunes stability (Figure~\ref{fig:exp1_diabetes}): OC4 (which prioritizes stability) achieves an LS value 0.045 better than any other configuration, at some cost to F1 and TPRD, trade-offs that \sys makes transparent. To our knowledge, \sys is the first ML system capable of tuning for model stability, and, built on \openbx with a standardized tuner API, it is easily extensible for advancing stability without sacrificing other dimensions. 

\textbf{In summary,} we answer \textbf{RQ1} affirmatively: \sys effectively optimizes ML pipelines under diverse multi-objective criteria, including fairness across binary and intersectional groups and model stability.

\subsection{Comparison to Other Systems}
\label{sec:comparison}

We compare \sys to state-of-the-art AutoML systems, \alpine, \askl, and \flaml, along two dimensions: performance under context-sensitive, multi-objective criteria (\textbf{RQ2.1}) and scalability with increasing computational resources (\textbf{RQ2.2}). All systems operate within the same search space and are allocated equal total computational resources, as described in Section~\ref{sec:exp_setup}.

\begin{table*}[t]
\footnotesize
\centering
\caption{Mean and standard deviation of the performance metrics for the pipeline selected by each system. Bold text marks the best value per metric, while gray shading shows the best \textit{average score} across multiple metrics using equal weighting.}
\label{tab:exp2}
\begin{tabular}{lccccccc}
\toprule
    \multirow{2}{*}{\textbf{System}} & \multicolumn{2}{c}{\textbf{Diabetes ($\approx$1K)}} & \multicolumn{2}{c}{\textbf{Folk Employment (15K)}} & \multicolumn{3}{c}{\textbf{Folk Public Coverage (50K)}} \\

\cmidrule(lr){2-3} \cmidrule(lr){4-5} \cmidrule(lr){6-8}
 & F1 & Label Stability & F1 & FNRD (sex \& race) & F1 & SRD (sex) & SRD (race) \\
 
\midrule
    autosklearn & 0.892 \scriptsize{$\pm$0.049} & 0.904 \scriptsize{$\pm$0.017} & 0.808 \scriptsize{$\pm$0.004} &  0.012 \scriptsize{$\pm$0.019} & \textbf{0.633 \scriptsize{$\pm$0.009}} & -0.033 \scriptsize{$\pm$0.010} &   0.177 \scriptsize{$\pm$0.008} \\
    
    alpine\_meadow & 0.915 \scriptsize{$\pm$0.034} & 0.889 \scriptsize{$\pm$0.011} & 0.809 \scriptsize{$\pm$0.007} &  0.014 \scriptsize{$\pm$0.023} & 0.628 \scriptsize{$\pm$0.006} & -0.033 \scriptsize{$\pm$0.008} &   0.178 \scriptsize{$\pm$0.011} \\

    flaml & 0.919 \scriptsize{$\pm$0.029} & 0.907 \scriptsize{$\pm$0.016} & \cellcolor{gray!20}0.802 \scriptsize{$\pm$0.005} & \cellcolor{gray!20}\textbf{-0.001 \scriptsize{$\pm$0.001}} & 0.588 \scriptsize{$\pm$0.039} & -0.037 \scriptsize{$\pm$0.015} &   0.094 \scriptsize{$\pm$0.056} \\
    
    virny\_flow & \cellcolor{gray!20}\textbf{0.929 \scriptsize{$\pm$0.023}} & \cellcolor{gray!20}\textbf{0.908 \scriptsize{$\pm$0.019}} & \cellcolor{gray!20}\textbf{0.811 \scriptsize{$\pm$0.005}} & \cellcolor{gray!20}-0.010 \scriptsize{$\pm$0.010} & \cellcolor{gray!20} 0.596 \scriptsize{$\pm$0.029} & \cellcolor{gray!20}\textbf{-0.011 \scriptsize{$\pm$0.007}} &   \cellcolor{gray!20}\textbf{0.068 \scriptsize{$\pm$0.029}} \\
\bottomrule
\end{tabular}
\end{table*}

\textbf{Performance under multi-objective criteria (RQ2.1).}  We evaluate performance on three datasets, \diabetes, \folkemp, and \folkpc, each associated with its own optimization criterion. Each system is run for 60 minutes per dataset and random seed. \sys and \flaml perform multi-objective optimization using equal weights for each metric, while \alpine and \askl optimize for F1 only, as they do not natively support multi-objective optimization.
Table~\ref{tab:exp2} reports the mean and standard deviation of performance metrics for the selected pipeline identified by each system across ten random seeds. Bold text indicates the best value for individual metrics, while gray shading highlights the best \textit{average score} across metrics using equal weighting. Overall, \sys achieves better or comparable performance to all three baselines across all datasets in terms of average score. On \diabetes, \sys clearly outperforms the baselines across all metrics. On \folkpc, \sys incurs a small decrease in F1 (0.037) while substantially improving fairness, particularly SRD for race (by 0.11), illustrating a favorable accuracy–fairness trade-off. On \folkemp, \sys matches the average score of \flaml, achieving a higher F1 score (by 0.009) at the cost of a slightly worse FNRD (by 0.009), which remains within acceptable fairness bounds.

\textbf{In summary}, context-sensitive, multi-objective optimization in \sys does not require sacrificing predictive performance: \sys achieves competitive or superior performance compared to all three baselines under identical resource constraints.

\looseness=-1
\textbf{Scalability and efficiency (RQ2.2).} We evaluate scalability on three datasets of increasing size: \heart (70K samples), \folkempbig (200K), and \folkempall (2.6M; the ACSEmployment task over all US states in 2018), varying the number of workers (one CPU core per worker) and measuring speedup. Since \alpine and \askl are limited to single-node execution, their evaluations cap at 32 workers; to ensure strict hardware consistency and isolate architectural differences from infrastructure-specific optimizations, we also evaluate \flaml on a single node (see \inappendix{Appendix~\ref{apdx:baselines}}). \sys supports distributed parallelism and is evaluated with up to 128 workers across four nodes. On \folkempall, we use 3 random seeds and a budget of 200 pipelines per run, and report speedup relative to \sys's 4-worker configuration, the smallest feasible at this scale; full results for \heart and \folkempbig are provided in \inappendix{Appendix~\ref{apdx:scalability}}.

Figure~\ref{fig:exp3_folk_emp_all_main} shows the results at the largest scale. \sys reaches a speedup of 30.7$\times$ at 128 workers, reducing optimization time from 40.1 hours (4 workers) to 1.3 hours, while \askl and \flaml plateau at 9.2$\times$ and 8.9$\times$, respectively, at their 32-worker cap---a 3.35$\times$ advantage for \sys beyond the point where single-node architectures stop scaling. \alpine could not complete this workload under the same resource constraints as the other systems: it exhausted available memory even with 512 GB per node (e.g., at 16 and 32 workers), our per-node hardware limit, and is therefore excluded. Crucially, scale does not degrade quality: across 4 to 128 workers, \sys's F1 remains within $[0.8208, 0.8226]$ and FNRD within $[-0.011, -0.007]$ (Figure~\ref{fig:exp3_folk_emp_all_main}, right), comparable to \flaml, the only baseline that also optimizes fairness, while \askl's FNRD ($\approx$0.032) is several times larger in absolute value. With fewer than 8 workers, \sys exhibits overhead from distributed coordination (Apache Kafka~\cite{kafka}), which impacts small-scale efficiency but is amortized at scale.

\begin{figure}[t!]
    \centering
    \begin{subfigure}{0.49\linewidth}
        \centering
        \includegraphics[width=\linewidth]{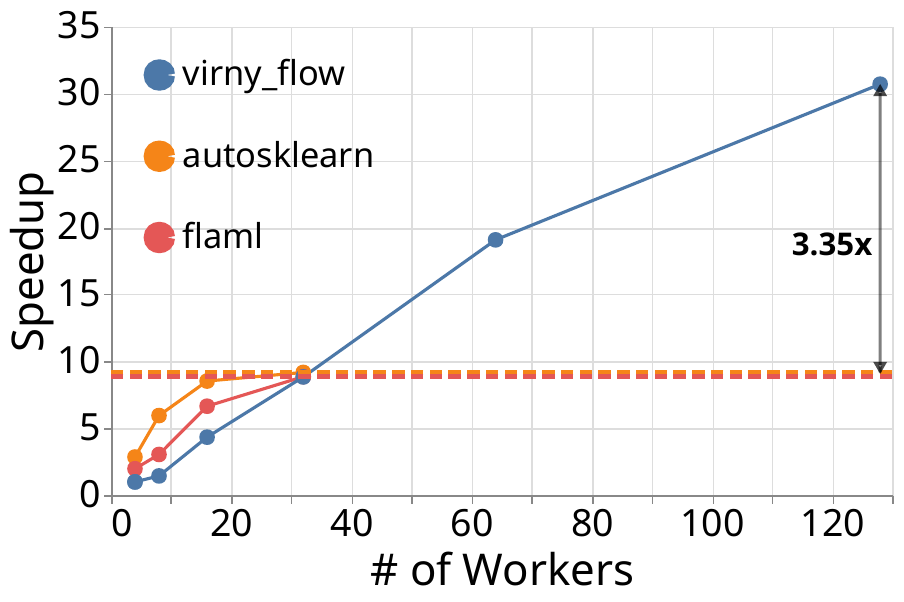}
        \label{fig:exp3_speedup_folk_emp_all_main}
    \end{subfigure}\hfill
    \begin{subfigure}{0.49\linewidth}
        \centering
        \includegraphics[width=\linewidth]{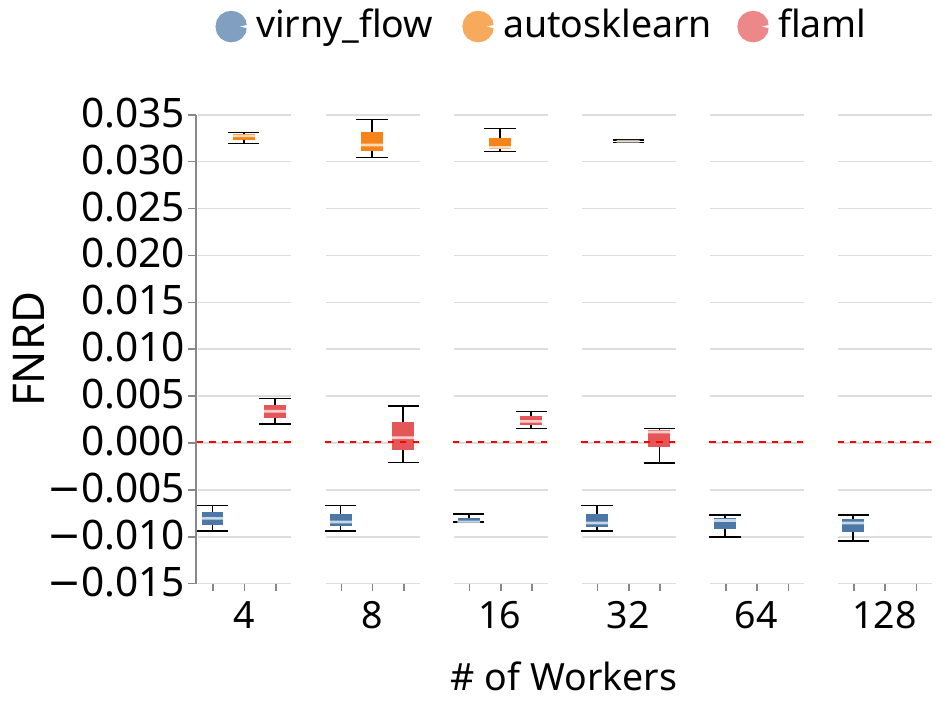}
        \label{fig:exp3_fnrd_folk_emp_all_main}
    \end{subfigure}
    \caption{Scalability on \folkempall ($\approx$2.6M rows): speedup (left) and FNRD (right) with varying numbers of workers. \alpine runs out of memory at this scale; \askl and \flaml do not scale beyond 32 workers. FNRD remains stable as \sys scales to 128 workers.}
    \label{fig:exp3_folk_emp_all_main}
\end{figure}

\textbf{In summary}, \sys demonstrates clear scalability advantages over existing AutoML systems, particularly in multi-node settings, while maintaining accuracy and fairness as parallelism increases. Additional scalability results are provided in \inappendix{Appendix~\ref{apdx:scalability}}.

\subsection{System Configuration Sensitivity}
\label{sec:exp4}

We examine the sensitivity of \sys to different configuration settings (\textbf{RQ3}). Results for varying the number of physical pipeline candidates ($k$) per logical pipeline selection (Section~\ref{sec:optimization}), which affects execution parallelism and the exploration–exploitation trade-off, together with additional sensitivity results, are reported in \inappendix{Tables~\ref{tab:sensitivity_pp} and~\ref{tab:sensitivity_halting} in Appendix~\ref{apdx:sensitivity}}. Experiments are conducted on two datasets using a single optimization criterion with equal weights for all component metrics. For \heart, the objective averages F1 and TNRD; for \folkpc, it averages F1, SRD for sex, and SRD for race. The search space matches that of Section~\ref{sec:comparison}, with two added fairness-enhancing interventions: Disparate Impact Remover~\cite{feldman2015certifying} and Adversarial Debiasing~\cite{zhang2018mitigating}. Each configuration evaluates up to 200 pipelines. We report the mean and std. of the composite objective score and runtime per run. \inappendixbig{Table~\ref{tab:sensitivity_pp}} shows a trade-off: increasing the number of physical pipeline candidates ($k$) reduces runtime but slightly degrades performance, with $k=4$ providing a strong balance. \inappendixbig{Table~\ref{tab:sensitivity_halting}} shows that varying training set fractions for pruning does not consistently reduce runtime, and small fractions can harm performance. For example, \{0.5, 1.0\} performs better on \heart, while \{0.75, 1.0\} yields a better trade-off on \folkpc.


\textbf{In summary,} in response to \textbf{RQ3}, \sys exposes meaningful trade-offs between pipeline performance and runtime, enabling informed configuration choices rather than reliance on fixed defaults.

\subsection{Case Study: Clinical Distribution Shift}
\label{sec:case_study_short}

\begin{table*}[t!]
    \centering
    \begin{minipage}[t]{0.45\linewidth}
        \centering
        \caption{Age cohorts and performance under distribution shift. Trained on 80\% of C (55--64 y.o.). F1 drops on younger cohorts; FNRD varies.}
        \label{tab:age_cohorts_exp1}
        \footnotesize 
        \begin{tabular}{lcccc}
            \toprule
            \textbf{Test cohort} & \textbf{\# Patients} & \textbf{Base} & \textbf{F1} & \textbf{FNRD} \\
            \midrule
            C (55--64 y.o.) & 31{,}255 & 0.609 & 0.764 & 0.002 \\
            B (45--54 y.o.) & 28{,}473 & 0.453 & 0.562 & -0.029 \\
            A (29--44 y.o.) & 10{,}272 & 0.298 & 0.199 & -0.006 \\
            \bottomrule
        \end{tabular}
    \end{minipage}
    \hfill
    \begin{minipage}[t]{0.50\linewidth}
        \centering
        \caption{Re-optimization on B+C (45--64 y.o.), evaluated on A (29--44 y.o.). Increasing FNRD weight improves the trade-off.}
        \label{tab:age_cohorts_exp3}
        \footnotesize 
        \begin{tabular}{llccc}
            \toprule
            \textbf{\shortstack{Weights}} & \textbf{Cohort} & \textbf{F1} & \textbf{FNRD} & \textbf{Score} \\
            \midrule
            \multirow{2}{*}{\textbf{F1} 0.5 / \textbf{FNRD} 0.5} 
                & B+C & 0.728 & -0.002 & 0.863 \\
                & A   & 0.674 & 0.044  & 0.815 \\
            \midrule
            \multirow{2}{*}{\textbf{F1} 0.25 / \textbf{FNRD} 0.75} 
                & B+C & 0.727 & 0.001 & 0.863 \\
                & A   & 0.676 & 0.016 & 0.830 \\
            \bottomrule
        \end{tabular}
    \end{minipage}
\end{table*}

To illustrate \sys in a deployment-relevant setting, we present a case study on cardiovascular disease (CVD) risk prediction using the \heart dataset. The model guides clinical screening decisions; patients are directly affected by missed or delayed diagnoses, making FNRD the primary fairness metric, since women are disproportionately underdiagnosed for CVDs~\cite{betai2024gender}. Accuracy must also remain robust under distribution shift, as clinical AI systems frequently degrade when deployed across patient populations with different age profiles~\cite{lee2023stable}.

\textbf{Distribution shift simulation.} We partition the data into three age cohorts based on clinically established cardiovascular risk categories~\cite{arnett2019}: Cohort A (low-risk, ages~$\leq\!44$), B (intermediate-risk, $45\!\leq$~ages~$\leq\!54$), and C (high-risk, ages~$\geq\!55$). The hospital initializes the screening pipeline on Cohort~C with equal F1/FNRD weights (0.5/0.5) to reflect an initial policy balancing overall
accuracy with error rate parity, using the same search space and hyperparameters as in Section~\ref{sec:exp4}. The selected pipeline is then evaluated on a held-out 20\% split of Cohort C and on the full Cohorts A and B, simulating deployment to younger populations. Dataset characteristics by age group and gender are provided in \inappendix{Appendix~\ref{sec:case_study} (Figure~\ref{fig:heart_stats})}.
Table~\ref{tab:age_cohorts_exp1} reports the impact of this shift: F1 drops by 20.2\% on Cohort B and by 56.5\% on Cohort~A relative to~C, while FNRD varies across cohorts, reflecting a prevalence-driven distribution shift.

\textbf{Distribution shift recovery.} Upon observing degradation on Cohorts A and~B, the clinical team incorporates intermediate-risk data and re-optimizes \sys using Cohort B+C as training data. Table~\ref{tab:age_cohorts_exp3} reports results on a 20\% test set of B+C and on the full Cohort A, where the composite Score $= 0.5{\cdot}\text{F1} + 0.5{\cdot}(1-|\text{FNRD}|)$ (range 0--1; higher is better). With the original 0.5/0.5 weights, \sys recovers 47.5\% of the
F1 degradation on Cohort~A while maintaining FNRD within an acceptable range.

The team then inspects the Pareto frontier for the 0.5/0.5 configuration via \sys's visualization interface\arxivonly{ (Figure~\ref{fig:pareto_frontier_exp2})}, observing that F1 is harder to improve than FNRD---a trade-off between predictive accuracy and clinically harmful false negatives. Rather than resolving this trade-off automatically, the team considers a 0.25/0.75 (F1/FNRD) configuration that prioritizes fairness while maintaining competitive accuracy. The selected pipeline (XGBoost with Adversarial Debiasing) achieves substantially lower FNRD on Cohort~A (0.016 vs.\ 0.044) with a marginal gain in F1, yielding a higher composite score (0.830 vs.\ 0.815). Importantly, both configurations lie on the Pareto frontier; the choice reflects a normative preference over trade-offs rather than a single optimal solution.

\textbf{In summary,} this case study demonstrates \sys's capacity to support deployment-relevant model development: grounding objectives in a clinical task, diagnosing and recovering from distribution shift through iterative re-optimization, and surfacing accuracy--fairness trade-offs for explicit human deliberation rather than automated resolution.

\section{User Study}
\label{sec:user_study}


We conducted an IRB-approved user study to evaluate whether \sys enables data scientists to reason about and deliberately navigate trade-offs among accuracy, fairness, stability, and other performance dimensions during model development. The study examines \first \textit{context sensitivity}, support for problem-specific, multi-objective pipeline design under domain constraints; and \second \textit{human centricity}, the extent to which human-in-the-loop mechanisms and visualizations support iterative refinement and informed decision-making. All study materials are available in the accompanying repository.\footnote{\url{https://github.com/DataResponsibly/virnyflow-user-study/}}

\textbf{Participants and protocol.} We recruited 10 participants (4 PhD students, 3 master's students, 1 postdoc, 1 bachelor's student, and 1 research staff member), representing the intended user population of \sys: data scientists responsible for defining evaluation protocols and design-time trade-offs. Participants reported high familiarity with Python, moderate familiarity with AutoML, and high to moderate familiarity with fairness and stability; they self-identified as women (3/10), men (6/10), and non-binary (1/10). Each participant completed a 90-minute single-group study consisting of a tutorial, two structured tasks based on \diabetes~\cite{diabetes_dataset}, and a post-study survey. The tasks required participants to interpret model performance across multiple dimensions, configure an experiment, and analyze outputs through interactive visualizations to determine whether the resulting trade-offs were acceptable in context; they were designed to elicit deliberation in settings with no single objectively correct solution.

\textbf{Results.} Participants achieved success rates of 89.3\% on the context-sensitivity tasks and 96.2\% on the human-centricity tasks (see \inappendix{Table~\ref{tab:user_study_results} in Appendix~\ref{apdx:user_study_additional}}), with average post-study ratings of 4.4 and 4.275, respectively, on a 5-point Likert scale, and an overall usability rating of 4.6. Beyond task completion, participants articulated, compared, and justified the relative importance of competing objectives and groups, treating evaluation protocol definition as a substantive design decision rather than a purely technical configuration step, and consistently described the visualization interface as supporting comparison and deliberation over alternative pipelines rather than merely presenting results. As one participant observed: \textit{``I had not previously thought about the problem in terms of a Pareto frontier, and I found this visualization very useful. [...] Overall, the visualizations are quite effective, and the hyperparameter importance view was particularly helpful.''} Participants perceived substantial time savings---most estimated that completing the same steps without \sys would take at least twice as long---and framed these savings as making it feasible to engage more deeply with multi-objective, fairness-aware model development. Suggested improvements included more comprehensive documentation, support for environments without Docker or Singularity, and additional explanation for some views; participants noted that the initial learning curve primarily reflected the inherent complexity of reasoning about multiple objectives rather than difficulties with basic interaction.

\textbf{In summary,} the user study provides evidence that \sys supports responsible model development as an iterative and deliberative process: participants reasoned about and navigated trade-offs among competing objectives, compared alternative pipelines, and refined configurations, while reporting high system usability.

\section{Conclusion}
\label{sec:conc}

This paper introduced \sys, a system that optimizes ML pipelines jointly for accuracy, fairness, and stability at scale, in which user-defined evaluation protocols drive optimization end to end, from Bayesian optimization of physical pipelines through bandit-based logical pipeline selection to multi-criterion pruning. Through experiments on six real-world datasets and an IRB-approved user study, we showed that context-sensitive, multi-objective optimization need not sacrifice performance or scale: \sys achieves competitive or superior results compared to state-of-the-art AutoML systems under identical resource constraints, scales to 128 workers across four nodes on datasets of up to 2.6M records while maintaining stable accuracy and fairness, and supports human-in-the-loop trade-off reasoning.

Much responsible-AI work focuses on auditing trained and deployed models, yet many consequential decisions are made earlier, during pipeline construction, metric selection, and optimization, and these decisions shape model behavior and downstream impacts~\cite{DBLP:journals/cacm/StoyanovichAHJS22,suresh2021framework,corbett2018measure}. By making evaluation protocols and optimization criteria explicit and revisable, \sys surfaces these choices before deployment, shifting the role of the system from producing a single ``best'' pipeline to supporting informed deliberation over alternatives~\cite{weerts2024can,shang2019democratizing}.

\textbf{Limitations and future work.} \sys targets supervised, fixed-structure pipelines over tabular data, and our evaluation is limited to binary classification, reflecting the lack of widely adopted context-sensitive fairness metrics in other settings; more complex dynamics such as temporal dependence, feedback loops, and online learning are beyond scope. Future work includes alternative multi-objective scoring beyond weighted sums~\cite{karl2023multi}, multi-objective meta-learning to warm-start optimization~\cite{ye2021multi}, refined pruning strategies, and participatory extensions that bring affected communities into evaluation protocol definition, the step that determines which objectives are optimized, and for whom. An extended discussion, with implications for education, governance, and organizational practice, is in \inappendix{Appendix~\ref{apdx:discussion}}.
\section*{Acknowledgments}
\label{sec:Acknowledgments}

Part of this work was conducted through the RAI for Ukraine program of the NYU Center for Responsible AI.
This work was supported in part by NSF Awards No. 2312930 and 2326193, and by the NYU IT High Performance Computing resources, services, and staff expertise.  

The authors used a large language model (Claude, Anthropic) to assist with restructuring and editing the text of this paper. All technical content, experiments, and analyses were conducted by the authors, who reviewed and verified all text and take full responsibility for the content of the paper.


\bibliographystyle{IEEEtran}
\bibliography{AnonymousSubmission/LaTeX/aaai2026_condensed.bib}

\ifarxiv
\clearpage
\appendices
\section{Extended Discussion}
\label{apdx:discussion}


This appendix expands on the discussion in Section~\ref{sec:conc}, restoring material condensed from the conference format.

\textbf{Design-time, not post-hoc.} Much RAI work focuses on auditing trained and deployed models. Yet many consequential decisions are made earlier, during pipeline construction, metric  selection, and optimization~\cite{DBLP:journals/cacm/StoyanovichAHJS22,mitchell2019model,passi2019problem,suresh2021framework}. These upstream choices encode normative assumptions that shape model behavior and downstream impacts~\cite{corbett2018measure,green2020algorithmic}; treating them as technical defaults rather than value judgments obscures accountability. By making evaluation protocols and optimization criteria explicit and revisable, \sys foregrounds these choices before deployment.

\textbf{From optimization to deliberation.} Most AutoML frameworks implicitly encode values, by fixing objectives, aggregating metrics, or prioritizing efficiency over transparency~\cite{passi2019problem,selbst2019fairness,corbett2018measure}. This work demonstrates that scalable, multi-objective optimization need not work this way: human oversight and interpretable trade-offs are compatible with, not in tension with, strong empirical performance~\cite{weerts2024can,robertson2024human}. The design-space framing shifts the role of the system from producing a ``best'' pipeline to supporting informed deliberation over alternatives~\cite{baratchi2024automated,shang2019democratizing}.

\textbf{Education and governance.} Benchmark-driven, single-objective training obscures the value-laden nature of metric selection~\cite{ijaied,fiesler2020ethics}; design-space systems make trade-offs concrete and can support responsible ML as a foundational engineering skill~\cite{stoyanovich_meta_eaai25,holstein2019improving,madaio2020codesign,stoyanovich_weareai_eaai25}. More broadly, preserved evaluation protocols, experiment histories, and comparative pipeline analyses can support internal review, documentation, and accountability workflows~\cite{mitchell2019model,crisan2022interactive}, aligning with emerging approaches that document not only model outcomes but the reasoning and trade-offs behind them~\cite{chmielinski2024clear,natmed_25,DBLP:journals/debu/StoyanovichH19}.

\textbf{Limitations.} We focus on supervised ML pipelines over tabular datasets, considering hundreds to thousands of pipeline variants with fixed structure. Extensions to neural architecture search, distributed training, unsupervised learning, and automated data acquisition are beyond scope. Our evaluation is limited to binary classification, reflecting the lack of widely adopted context-sensitive fairness metrics in other settings. While we introduce a deployment-oriented case study with task-grounded objectives and distribution shift as initial evidence, our evaluation does not yet capture more complex dynamics such as temporal dependence, feedback loops, or online learning. We do not study interactions between evaluation protocols and optimizers, nor end-to-end use across diverse tasks.

\textbf{Future work} includes exploring alternative scoring methods for multi-objective optimization beyond weighted sums, such as Tchebycheff and $\epsilon$-constraint approaches~\cite{karl2023multi}, incorporating multi-objective meta-learning to warm-start optimization~\cite{ye2021multi}, and refining pruning strategies to further improve resource efficiency while maintaining transparency and usability. On the socio-technical side, future work should examine how \sys is adopted and adapted within organizational and institutional contexts, and explore participatory extensions that bring affected communities into the evaluation protocol definition process, the step where normative choices about which objectives matter, and for whom, are most consequential.

\section{Cardiovascular Disease Case Study}
\label{sec:case_study}

\begin{figure}[tb]
  \centering
  \includegraphics[width=\linewidth]{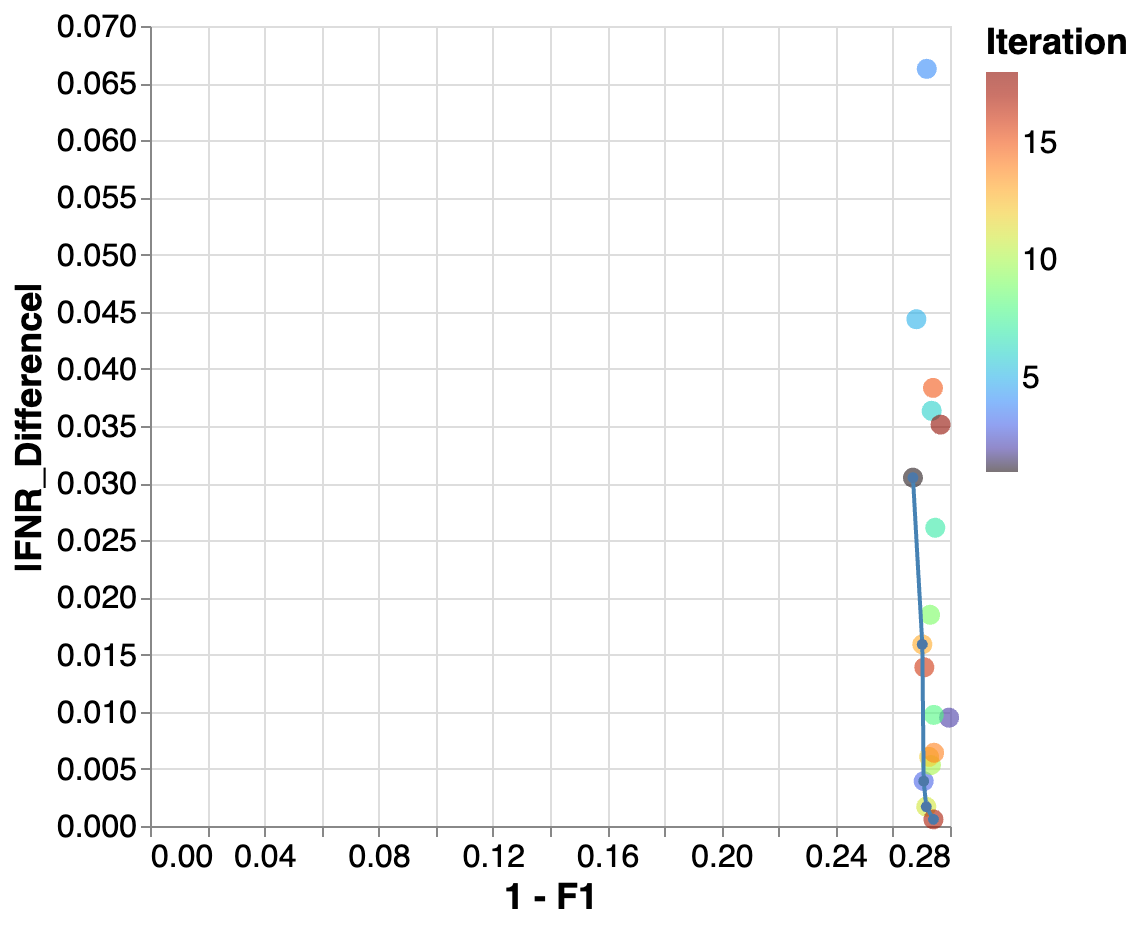}
  \caption{Pareto frontier from VirnyFlow's interface for the 0.5/0.5 weight configuration from Table~\ref{tab:age_cohorts_exp3}. Both objectives are cast as minimization targets: F1 is transformed to (1 - F1) and FNRD is taken in absolute value. Each point represents a pipeline evaluated during optimization, with color indicating the iteration. The optimizer quickly drives $|FNRD|$ near zero, while (1 - F1) remains concentrated around 0.27-0.28, indicating that accuracy is the harder objective to improve in the \heart dataset.}
  \label{fig:pareto_frontier_exp2}
\end{figure}

\begin{figure*}[t!]
    \centering

    \begin{subfigure}{0.47\linewidth}
        \centering
        \includegraphics[width=\linewidth]{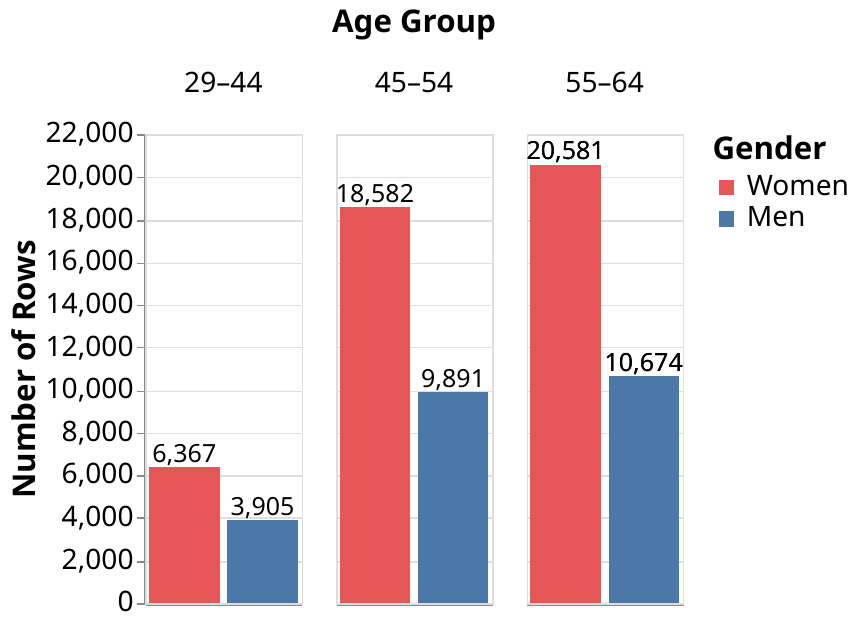}
        \caption{Number of rows per age group and gender}
        \label{fig:heart_proportion}
    \end{subfigure}
    \hspace{2cm}
    \begin{subfigure}{0.28\linewidth}
        \centering
        \includegraphics[width=\linewidth]{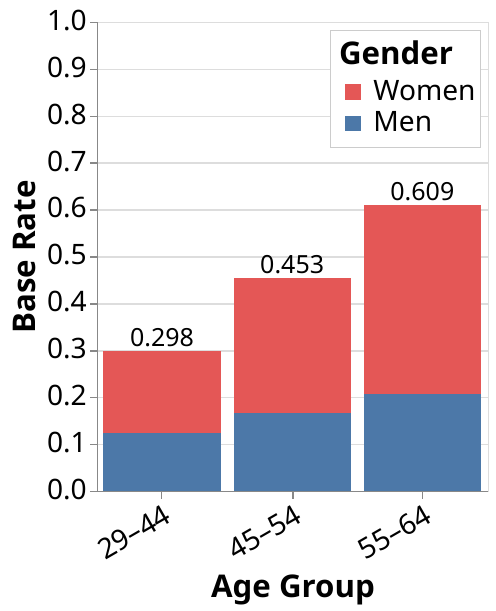}
        \caption{Base rates per age group and gender}
        \label{fig:heart_base_rate}
    \end{subfigure}

    \caption{Dataset characteristics of the \heart dataset. (a) Row counts per age group and gender subgroup, showing consistent female overrepresentation across all cohorts. (b) Positive class base rates per subgroup, revealing a pronounced age-driven prevalence shift that introduces natural distribution shift across groups.}
    \label{fig:heart_stats}
\end{figure*}

This appendix provides additional details for the case study introduced in Section~\ref{sec:case_study_short}, including clinical background, dataset characteristics, and supporting figures.

\textbf{Clinical background.}
Cardiovascular diseases (CVDs) are the leading cause of death worldwide, accounting for about 19.2 million deaths in 2023, or roughly one-third of all global fatalities~\cite{global2025global}. Around 80\% of this burden is linked to modifiable risk factors such as hypertension, obesity, and smoking, highlighting the importance of early screening~\cite{global2025global,oude2023health}. Traditional risk prediction models often fail to capture complex interactions among these factors, motivating the use of ML methods~\cite{banerjee2025systematic}.

As discussed in Section~\ref{sec:case_study_short}, clinicians use the model to guide screening decisions, while patients are directly affected by missed or delayed diagnoses. This introduces two core challenges. First, the screening pipeline must balance overall predictive performance with error rate parity across gender, as women are often underdiagnosed and undertreated for CVDs~\cite{betai2024gender, al2024gender}. False negatives are particularly harmful because missed diagnoses delay care, motivating the use of FNRD as the primary fairness metric~\cite{mihan2024artificial}. At the same time, accuracy remains important to avoid overburdening limited diagnostic resources with false positives. Second, the model must remain robust under distribution shift, as clinical AI systems often degrade when patient populations change~\cite{lee2023stable}. For example, a model trained on older, high-risk patients (age $\geq$ 55) may encounter different distributions of features such as blood pressure, cholesterol, and weight when a hospital expands its screening to younger populations~\cite{lee2023stable, subasri2025detecting}.

\textbf{Dataset characteristics.} Figure~\ref{fig:heart_stats} shows the \heart dataset breakdown by age cohort and gender, complementing Table~\ref{tab:age_cohorts_exp1} in Section~\ref{sec:case_study_short}. The monotonic increase in positive class prevalence from Cohort~A (low-risk, ages $\leq$ 44) through Cohort~B (intermediate-risk, 45 $\leq$ ages $\leq$ 54) to Cohort~C (high-risk, ages $\geq$ 55) supports the clinical relevance of the age thresholds~\cite{arnett2019, score22021score2}, while the consistent female overrepresentation across cohorts illustrates the demographic context in which fairness across gender must be evaluated.

\textbf{Pareto frontier inspection.} Figure~\ref{fig:pareto_frontier_exp2} shows the Pareto frontier surfaced by \sys's visualization interface (Appendix~\ref{apdx:viz_interfaces}) for the 0.5/0.5 weight configuration on Cohort~B+C. The optimizer quickly drives $|FNRD|$ near zero, while (1 - F1) remains concentrated around 0.27--0.28, indicating that F1 is the harder objective to improve in the \heart dataset. Observing this asymmetry---that fairness was already being driven near zero while accuracy plateaued---the clinical team reasoned that shifting weight toward FNRD would cost little in accuracy while further reducing clinically harmful false negatives, motivating the choice of an alternative 0.25/0.75 (F1/FNRD) configuration as described in Section~\ref{sec:case_study_short}. Importantly, weights represent policy choices rather than parameters to be optimized: both the 0.5/0.5 and 0.25/0.75 configurations lie on the Pareto frontier, and the selection between them reflects a normative preference over trade-offs rather than the resolution of a single optimization problem. This pattern illustrates how \sys supports deliberative, design-time engagement with accuracy--fairness trade-offs rather than automated resolution.

\section{Conceptual Map of the VirnyFlow Search Space}
\label{apdx:search_space}

\begin{table*}[t!]
\caption{Pipeline components and optimization methods available in \sys.}
\label{tab:search_space}
\centering
\footnotesize
\setlength{\tabcolsep}{6pt}
\renewcommand{\arraystretch}{1.0}
\begin{tabularx}{\textwidth}{X X X X}
\toprule
\textbf{Optimizers} & \textbf{Null Imputers} & \textbf{Fairness Interventions} & \textbf{ML Models} \\
\midrule
\parbox[t]{\linewidth}{%
\textit{Surrogate models}~\cite{li2021openbox}:\\
- Gaussian Process\\
- Probabilistic Random Forest\\[4pt]

\textit{Acquisition functions}~\cite{li2021openbox}:\\
- Expected Improvement with / without Constraints\\
- Expected Hypervolume Improvement with / without Constraints\\
- Max-value Entropy Search for Multi-Objective with / without Constraints\\ [4pt]

\textit{Acq. function optimizers}~\cite{li2021openbox}:\\
- Interleaved Local \& Random Search\\
- Random Search \& L-BFGS-B}
&
\parbox[t]{\linewidth}{%
- deletion\\
- median-mode\\
- median-dummy\\
- missForest~\cite{stekhoven2012missforest}\\
- k-means clustering~\cite{gajawada2012missing}\\
- datawig~\cite{biessmann2019datawig}\\
- automl~\cite{jager2021benchmark}}
&
\parbox[t]{\linewidth}{%
\textit{Pre-processing}:\\
- Disparate Impact Remover~\cite{feldman2015certifying}\\
- Learning Fair Representations~\cite{zemel2013learning}\\[4pt]

\textit{In-processing}:\\
- Adversarial Debiasing \cite{zhang2018mitigating}\\
- Exponentiated Gradient Reduction~\cite{agarwal2018reductions}\\[4pt]

\textit{Post-processing}:\\
- Equalized Odds Postprocessing~\cite{hardt_EOP2016}\\
- Reject Option Classification~\cite{kamiran2012decision}}
&
\parbox[t]{\linewidth}{%
\textit{Sklearn models}~\cite{scikit-learn}:\\
- Decision Tree\\
- Linear Regression\\
- Logistic Regression\\
- SVM\\
- K-Neighbors\\
- Random Forest\\
- MLP\\
- XGBoost\\
- LGBM\\ [4pt]

\textit{Pytorch-tabular models}~\cite{pytorch-tabular}:\\
- GANDALF\\
- DANet\\
- MDN\\
- NODE\\
- FT Transformer\\
- GATE}
\\
\bottomrule
\end{tabularx}
\end{table*}

To provide a more formal characterization of the \sys search space, we outline the system’s supported pipeline configurations, optimization capabilities, and objective taxonomy.

\paragraph{Pipeline Configuration Space.} Each pipeline in \sys is constructed by composing components across multiple stages, including imputation, fairness intervention, and model training. These choices define the space of logical pipelines to be explored during optimization. A detailed listing of all supported pipeline components, grouped by stage, is presented in Table~\ref{tab:search_space}. Additionally, each pipeline component has an associated hyperparameter search space, which is used to instantiate physical pipelines. These hyperparameter sets are defined in the \sys codebase.\footnote{\url{https://github.com/denysgerasymuk799/virny-flow/blob/main/virny_flow/configs/component_configs.py}} Importantly, users may also extend this configuration space by providing custom components, enabling experimentation with novel preprocessing methods, fairness interventions, or model architectures tailored to their specific domain.

\paragraph{Optimization Methods.} \sys uses Bayesian optimization to efficiently search over pipeline configurations. Built on top of \texttt{OpenBox}~\cite{li2021openbox}, our system supports multiple surrogate models, acquisition functions (including constraint-aware variants), and optimizers that balance exploration and exploitation. These methods are summarized in Table~\ref{tab:search_space}, alongside their corresponding categories. Together, they enable scalable, context-sensitive tuning of ML pipelines aligned with user-defined goals. In addition to the built-in options, users may also provide custom optimizers that conform to the \texttt{OpenBox} API, allowing them to experiment with novel search strategies and tailor the optimization process to their application-specific needs.

\paragraph{Objective Taxonomy.} \sys supports optimization over a rich set of performance metrics, organized into three high-level dimensions:

\begin{itemize}[leftmargin=*]
  \item \textbf{Correctness:}
  \begin{itemize}
    \item Standard predictive metrics such as \textit{F1 score, accuracy, precision (PPV)}, and \textit{recall}.
    \item Confusion-matrix-derived metrics including \textit{TPR, TNR, FPR}, and \textit{FNR}. These reflect different aspects of classification performance such as sensitivity (TPR), specificity (TNR), and types of errors.
  \end{itemize}

  \item \textbf{Stability \& Uncertainty:}
  \begin{itemize}
    \item \textit{Label Stability}~\cite{darling2018toward}: Measures the consistency of model’s predictions when trained on different bootstrapped samples of the dataset. Higher stability implies greater robustness to sampling variability.
    \item \textit{Epistemic Uncertainty}~\cite{tahir2023fairness}: Represents uncertainty arising from limited knowledge about the best model parameters. It reflects the model’s confidence and can be reduced with more data or better modeling assumptions.
    \item \textit{Aleatoric Uncertainty}~\cite{tahir2023fairness}: Captures irreducible noise inherent in the input data. This uncertainty remains even with unlimited training data and often reflects ambiguity or label noise.
  \end{itemize}
  
  \item \textbf{Fairness:}
  \begin{itemize}
    \item \textit{Equalized odds (TPRD, TNRD, FPRD, FNRD)}~\cite{hardt_EOP2016}: Assess disparities in classification error rates and correct prediction rates across groups. These include sensitivity (TPRD), specificity (TNRD), false positive/negative rate differences, and highlight whether subgroups are disproportionately advantaged or harmed.
    \item \textit{Statistical Parity Difference (a.k.a. Selection Rate Difference)}~\cite{cv2010}: Measures the difference in the proportion of positive predictions between groups, regardless of correctness. A value close to zero indicates equal treatment.
    \item \textit{Disparate Impact}~\cite{chouldechova2017fair}: Defined as the ratio of positive prediction rates between groups. It emphasizes proportional fairness and is commonly used in legal and regulatory contexts.
    \item \textit{Label Stability Difference}~\cite{arifkhan2023stability}: Captures disparities in prediction consistency across groups. A higher difference indicates that one group experiences more variability in outcomes.
    \item \textit{Epistemic Uncertainty Difference}~\cite{arifkhan2023stability}: Measures whether uncertainty due to limited knowledge varies between groups, which may reflect biased model confidence.
    \item \textit{Aleatoric Uncertainty Difference}~\cite{arifkhan2023stability}: Indicates group-level differences in irreducible data noise, helping identify whether certain populations are systematically harder to model.
    \item All fairness metrics can be computed over binary or intersectional subgroups to capture nuanced disparities that may otherwise be overlooked.
  \end{itemize}
\end{itemize}

\section{Additional Details on Bayesian Optimization}
\label{sec:bo}

Bayesian optimization (BO)~\cite{frazier2018bayesian, archetti2019bayesian, garnett2023bayesian}  is a sample-efficient strategy for exploring complex, high-dimensional search spaces---such as those encountered in ML pipeline tuning. BO is particularly well-suited for optimizing black-box, expensive, and multi-extremal objective functions~\cite{candelieri2024fair, li2021openbox}, providing the necessary flexibility to integrate stability into pipeline tuning. Its effectiveness is further supported by recent surveys~\cite{pfisterer2022yahpo} and benchmarks~\cite{karl2023multi}, which highlight its sample efficiency for tabular datasets.

BO relies on building a probabilistic approximation of the objective function $\mathbf{f}$, often called a \textit{surrogate model}, based on previously evaluated pipeline configurations. BO proceeds iteratively, balancing exploration and exploitation by incorporating two key components: (1) a \textit{surrogate model} that captures our current belief about the performance landscape, and (2) an \textit{acquisition function} that suggests the next most promising configuration to evaluate.

The \textit{surrogate model} learns a posterior distribution over $\mathbf{f}$ by updating a prior belief using observations from prior evaluations. This posterior captures both the expected performance and the uncertainty associated with different regions of the search space~\cite{Jones1998,7352306}. The \textit{acquisition function} then leverages this posterior, typically combining its mean and variance, to prioritize candidate configurations that are either likely to perform well or lie in high-uncertainty regions, thus enabling efficient and informed exploration during the tuning process.

BO extends naturally to the multi-objective setting (MOBO)~\cite{candelieri2024fair}, where the goal is to efficiently approximate the \textit{Pareto front} of trade-offs among multiple, often conflicting, objectives. MOBO maintains a probabilistic \textit{surrogate model} for each objective, assuming independence, and uses an \textit{acquisition function} to balance exploration and exploitation in selecting new configurations to evaluate. Since objective values are typically expensive to query and only known at specific points, MOBO emphasizes sample efficiency. Common strategies include \textit{scalarization}~\cite{pmlr-v115-paria20a, pmlr-v119-zhang20i}, which reduces multiple objectives to a single weighted sum; \textit{hypervolume-based} methods, such as Expected Hypervolume Improvement (EHVI)~\cite{emmerich2006single}, which aim to maximize the dominated region under the Pareto front; and \textit{information-theoretic} approaches~\cite{belakaria2019max, pmlr-v119-suzuki20a, Belakaria2020}, which reduce uncertainty about the front. These techniques guide the search toward diverse, high-performing solutions with minimal evaluations.

Recent multi-objective Bayesian optimization (MOBO) methods~\cite{perrone2021fair, candelieri2024fair} have demonstrated competitive performance in incorporating fairness into the optimization process compared to other multi-objective hyperparameter optimization (HPO) techniques. Another advantage of BO is the availability of robust and efficient software frameworks~\cite{li2021openbox, balandat2020botorch, ax2025}, which provide standardized APIs to interact with different optimizers.

Multi-objective optimization (MO), unlike single-objective optimization,  does not yield a unique optimal solution. Instead, it produces the \textit{Pareto front} of dominant solutions, each representing a different trade-off among the objectives.  \sys incrementally approximates the Pareto front during pipeline optimization.

\section{Additional Implementation Details}
\label{apdx:execution}


\begin{algorithm}[tb]
\caption{Next Logical Plan (NLP)}
\label{alg:next_logical_plan}
\begin{algorithmic}[1]
\Require Problem $\mathcal{P}$, dataset $\mathcal{D}$, exploration factor $\beta$, risk factor $\theta$
\Ensure Next \textit{logical pipeline}
\If{$\text{rand}() < \beta$} \Comment{Exploitation}
    \State Compute $\mu_k$, $\delta_k$, and $c_k$ for each \textit{logical pipeline} $k$ using the history
    \State $\text{LogicalPlan} \gets$ select a \textit{logical pipeline} $k$ with probability proportional to $\mu_k + \frac{\theta}{c_k} \cdot \delta_k$
\Else \Comment{Exploration}
    \State $\text{LogicalPlan} \gets$ random unseen \textit{logical pipeline}
\EndIf
\State \Return $\text{LogicalPlan}$
\end{algorithmic}
\end{algorithm} 

Algorithm~\ref{alg:next_logical_plan}, \textit{Next Logical Plan} (NLP), details the bandit-based selection of \lps introduced in Section~\ref{sec:optimization}: with probability $\beta$ the currently most promising \lp is exploited based on the scoring model of Equation~\ref{eq:score}, and otherwise a random unseen \lp is explored.

Similar to \alpine~\cite{shang2019democratizing}, our \texttt{Task Queue} has a limited size of $m$, meaning that tasks are continuously added by the \texttt{Task Generator} until this limit is reached. This mechanism ensures that workers always have enough preloaded tasks available once they complete their current assignments. Additionally, the \textit{BO-advisor} generates not just one \textit{suggestion} but $k$ \textit{suggestions}, introducing random candidates to mitigate the risk of getting stuck in a local optimum. Thus, when $k$ slots become available in the queue, the next promising \lp is selected, and $k$ \pps are added to the queue. According to \alpine, this approach is effective under the assumption that the number of workers $w$ is significantly larger than $k$, i.e., $w \gg k$, and that the queue size $m$ is greater than $w$.

To further optimize performance, \sys minimizes the number of queries to the external database and indexes all database tables to accelerate retrieval.

\section{Additional Experimental Details}

\subsection{Datasets and Tasks}
\label{apdx:datasets}

Tables~\ref{tab:diabetes-rates}-\ref{tab:folk-emp-all-rates} report the demographic composition of all datasets, specifically, the proportions and base rates of each protected group. The datasets are described in the following paragraphs.

\emph{\diabetes}\footnote{\url{https://www.kaggle.com/datasets/tigganeha4/diabetes-dataset-2019}}~\cite{diabetes_dataset} was collected in India using a questionnaire comprising 18 questions covering aspects of health, lifestyle, and family background. It includes responses from 952 individuals, each described by 17 attributes --- 13 categorical and 4 numerical --- as well as a binary target variable indicating diabetes status. In this dataset, the sensitive attribute is \textit{sex}, with “female” identified as the disadvantaged group.

\begin{table*}[h!]
    \caption{Proportions and Base Rates for \diabetes ($\approx$1K).}
    \centering
    \footnotesize
    \begin{tabular}{|c|c|c|c|}
    \hline
     & overall & gender\_priv & gender\_dis \\ 
    \hline
    Proportions & 1.0 & 0.621 & 0.379 \\ 
    \hline
    Base Rates & 0.291 & 0.272 & 0.321 \\ 
    \hline
    \end{tabular}
    \label{tab:diabetes-rates}
\end{table*}

\begin{table*}[h!]
\caption{Proportions and Base Rates for \folkemp (15K).}
\centering
\footnotesize
    \begin{tabular}{|c|c|c|c|c|c|c|c|}
    \hline
     & overall & sex\_priv & sex\_dis & race\_priv & race\_dis & sex\&race\_priv & sex\&race\_dis \\ 
    \hline
    Proportions & 1.0  & 0.487 & 0.513 & 0.627 & 0.373 & 0.804 & 0.196 \\ 
    \hline
    Base Rates & 0.571 & 0.621 & 0.524 & 0.563 & 0.586 & 0.578 & 0.544 \\ 
    \hline
    \end{tabular}
\label{tab:folk-emp-rates}
\end{table*}

\begin{table*}[h!]
\caption{Proportions and Base Rates for \folkpc (50K).}
\centering
\footnotesize
    \begin{tabular}{|c|c|c|c|c|c|c|c|}
    \hline
     & overall & sex\_priv & sex\_dis & race\_priv & race\_dis & sex\&race\_priv & sex\&race\_dis \\ 
    \hline
    Proportions & 1.0  & 0.436 & 0.564 & 0.625 & 0.375 & 0.794 & 0.206 \\ 
    \hline
    Base Rates & 0.399 & 0.414 & 0.388 & 0.35 & 0.482 & 0.376 & 0.49 \\ 
    \hline
    \end{tabular}
\label{tab:folk-pubcov-rates}
\end{table*}

\begin{table*}[h!]
    \caption{Proportions and Base Rates for \heart (70K).}
    \centering
    \footnotesize
    \begin{tabular}{|c|c|c|c|}
    \hline
     & overall & gender\_priv & gender\_dis \\ 
    \hline
    Proportions & 1.0 & 0.35 & 0.65 \\ 
    \hline
    Base Rates & 0.5 & 0.505 & 0.497 \\ 
    \hline
    \end{tabular}
    \label{tab:heart-rates}
\end{table*}

\begin{table*}[h!]
\caption{Proportions and Base Rates for \folkempbig (200K).}
\centering
\footnotesize
    \begin{tabular}{|c|c|c|c|c|c|c|c|}
    \hline
     & overall & sex\_priv & sex\_dis & race\_priv & race\_dis & sex\&race\_priv & sex\&race\_dis \\ 
    \hline
    Proportions & 1.0  & 0.49 & 0.51 & 0.625 & 0.375 & 0.806 & 0.194 \\ 
    \hline
    Base Rates & 0.569 & 0.616 & 0.524 & 0.562 & 0.58 & 0.576 & 0.54 \\ 
    \hline
    \end{tabular}
\label{tab:folk-emp-big-rates}
\end{table*}

\begin{table*}[h!]
\caption{Proportions and Base Rates for \folkempall ($\approx$2.6M).}
\centering
\footnotesize
    \begin{tabular}{|c|c|c|c|c|c|c|c|}
    \hline
     & overall & sex\_priv & sex\_dis & race\_priv & race\_dis & sex\&race\_priv & sex\&race\_dis \\ 
    \hline
    Proportions & 1.0  & 0.486 & 0.514 & 0.779 & 0.221 & 0.885 & 0.115 \\ 
    \hline
    Base Rates & 0.566 & 0.606 & 0.528 & 0.569 & 0.556 & 0.57 & 0.538 \\ 
    \hline
    \end{tabular}
\label{tab:folk-emp-all-rates}
\end{table*}

Folktables~\cite{folktables_ding2021} is another popular fairness dataset derived from US Census data from all 50 states between 2014-2018. The dataset has several associated tasks, of which we selected two: (i) ACSPublicCoverage (\folkpc) is a binary classification task to predict whether a low-income individual, not eligible for Medicare, has coverage from public health insurance. The dataset contains 19 features (17 categorical, 2 numerical) including disability, employment status, total income, and nativity. We use data from New York from 2018, subsampled to 50k rows. (ii) ACSEmployment (\folkemp, \folkempbig, and \folkempall) is a binary classification task to predict whether an individual is employed, from 16 features (15 categorical, 1 numerical) including educational attainment, employment status of parent, military status, and nativity. We use data from California from 2018, subsampled to 15k and 200k for \folkemp and \folkempbig, respectively. \folkempall uses the same task and features, but covers all US states in 2018 without subsampling, yielding 2,590,315 rows after applying the standard ACSEmployment filters. In both tasks, \textit{sex} and \textit{race} are the sensitive attributes, with ``female'' and ``non-White'' as the disadvantaged groups.

\emph{\heart}\footnote{\url{https://www.kaggle.com/datasets/sulianova/cardiovascular-disease-dataset}} contains medical measurements related to cardiovascular conditions, covering 70,000 individuals. Each record includes 11 attributes --- 6 categorical and 5 numerical --- such as age, height, weight, and blood pressure, along with a binary target variable indicating the presence of heart disease. In this dataset, the sensitive attribute is \textit{sex}, with “female” considered the disadvantaged group.

\subsection{Model Types}
\label{apdx:model_types}

We evaluate predictive performance of 6 ML models in our experiments: (i) decision tree (\texttt{dt$\_$clf}) with a tuned maximum tree depth, minimum samples at a leaf node, number of features used to decide the best split, and criteria to measure the quality of a split; (ii) logistic regression (\texttt{lr$\_$clf}) with tuned regularization penalty, regularization strength, and optimization algorithm; (iii) light gradient boosted machine (\texttt{lgbm$\_$clf)} with tuned number of boosted trees, maximum tree depth, maximum tree leaves, and minimum number of samples in a leaf; (iv) random forest (\texttt{rf$\_$clf}) with a tuned number of trees, maximum tree depth, minimum samples required to split a node, and minimum samples at a leaf node; (v) extreme gradient boosting trees (\texttt{xgb$\_$clf}) with a tuned tree depth, learning rate, number of boosting rounds, subsample ratio of the training instances, and minimum sum of instance weight needed in a child node; (vi) a deep table-learning method called GANDALF~\cite{joseph2022gandalf} (\texttt{gandalf$\_$clf)} with a tuned learning rate, number of layers in the feature abstraction layer, dropout rate for the feature abstraction layer, and initial percentage of features to be selected in each Gated Feature Learning Unit (GFLU) stage. Search grids of hyperparameters for all models are defined in our codebase.

\subsection{Evaluation Metrics}
\label{apdx:exp:metrics}

To assess stability, we report average \emph{Label Stability (LS)}~\cite{darling2018toward, arifkhan2023stability} over the test set. For binary classification, LS = $\frac{|B_+ - B_-|}{B}$, where $B_+$ and $B_-$ are the counts of positive and negative predictions across $B=50$ bootstrapped models (80\% training fraction in our experiments).

To assess model fairness, we report error disparity metrics based on group-specific error rates, namely \emph{True Positive Rate Difference (TPRD)}, \emph{True Negative Rate Difference (TNRD)}, \emph{False Negative Rate Difference (FNRD)}, and \emph{Selection Rate Difference (SRD)}:

$$ TPRD = \frac{TP_{dis}}{TP_{dis}+FN_{dis}} - \frac{TP_{priv}}{TP_{priv}+FN_{priv}}$$

$$ TNRD = \frac{TN_{dis}}{TN_{dis}+FP_{dis}} - \frac{TN_{priv}}{TN_{priv}+FP_{priv}}$$

$$ FNRD = \frac{FN_{dis}}{TP_{dis}+FN_{dis}} - \frac{FN_{priv}}{TP_{priv}+FN_{priv}}$$

$$ SRD = \frac{TP_{dis}+FP_{dis}}{N_{dis}} - \frac{TP_{priv}+FP_{priv}}{N_{priv}}$$

\begin{table*}[h!]
\centering
\caption{Sensitivity of \sys to the training set fractions for pruning. Gray shadowing highlights the most optimal setting in terms of performance and efficiency.}
\label{tab:sensitivity_halting}
\begin{tabular}{lcccc}
\toprule
    \multirow{2}{*}{\textbf{Pruning}} & \multicolumn{2}{c}{\textbf{Heart (70K)}} & \multicolumn{2}{c}{\textbf{Folk Pub. Cov. (50K)}} \\

    \cmidrule(lr){2-3} \cmidrule(lr){4-5} 
    & Score & Runtime & Score & Runtime \\
    
\midrule
     \{1.0\} & 86.35 \scriptsize{$\pm$0.17} &    653 \scriptsize{$\pm$46} & 82.77 \scriptsize{$\pm$0.29} &  782 \scriptsize{$\pm$492} \\
         
    \{0.25, 1.0\} & 86.33 \scriptsize{$\pm$0.20} &   939 \scriptsize{$\pm$632} & 82.64 \scriptsize{$\pm$0.31} &  765 \scriptsize{$\pm$146} \\
    
    \{0.5, 1.0\} & \cellcolor{gray!20} 86.41 \scriptsize{$\pm$0.14} & \cellcolor{gray!20} 651 \scriptsize{$\pm$42} & 82.67 \scriptsize{$\pm$0.27} &   682 \scriptsize{$\pm$70} \\
     
    \{0.75, 1.0\} & 86.35 \scriptsize{$\pm$0.20} &   795 \scriptsize{$\pm$166} & \cellcolor{gray!20} 82.78 \scriptsize{$\pm$0.44} &  \cellcolor{gray!20} 789 \scriptsize{$\pm$120} \\
        
    \{0.25, 0.5, 1.0\} & 86.18 \scriptsize{$\pm$0.60} &   815 \scriptsize{$\pm$142} & 82.91 \scriptsize{$\pm$0.43} &  970 \scriptsize{$\pm$205} \\
    
    \{0.5, 0.75, 1.0\} & 86.36 \scriptsize{$\pm$0.08} &   766 \scriptsize{$\pm$111} & 82.74 \scriptsize{$\pm$0.33} &  871 \scriptsize{$\pm$205} \\
    
    \{0.1, 0.25, 0.5, 1.0\} & 86.28 \scriptsize{$\pm$0.30} &    850 \scriptsize{$\pm$75} & 82.84 \scriptsize{$\pm$0.44} & 1104 \scriptsize{$\pm$166} \\
    
    \{0.1, 0.5, 0.75, 1.0\} & 86.45 \scriptsize{$\pm$0.35} & 1403 \scriptsize{$\pm$1023} & 82.49 \scriptsize{$\pm$0.43} &  1007 \scriptsize{$\pm$81} \\
\bottomrule
\end{tabular}
\end{table*}

\subsection{Baselines}
\label{apdx:baselines}

We compare our system with three state-of-the-art, broad-spectrum AutoML baselines: \alpine~\cite{shang2019democratizing}, \texttt{auto-sklearn}~\cite{feurer2022auto}, and \flaml~\cite{wang2021flaml}, which were highlighted in recent benchmarks~\cite{neutatz2024automl, gijsbers2024amlb} and surveys~\cite{baratchi2024automated, barbudo2023eight}. Below, we briefly describe each baseline.

\begin{itemize}
    \item \alpine (version 1.0.1) is an interactive AutoML tool that won the DARPA D3M competition in April 2019. It applies concepts from query optimization and introduces novel pipeline selection and pruning strategies using cost-based multi-armed bandits and Bayesian optimization (BO).

    \item \askl (version 0.15.0) is a leading open-source AutoML system that tackles the CASH problem using Scikit-learn algorithms~\cite{scikit-learn}, incorporating BO, successive halving, ensembling, and meta-learning. It won the second ChaLearn AutoML challenge~\cite{guyon2019analysis}.

    \item \flaml (version 2.3.6) is a state-of-the-art AutoML system that prioritizes computational efficiency and flexible optimization; the MLSys’21 study introducing FLAML shows that it consistently achieves competitive or superior performance compared to other leading AutoML frameworks under equal or substantially smaller computational budgets~\cite{wang2021flaml}. FLAML supports multi-objective optimization and custom metrics and is integrated into Microsoft Fabric Data Science, making it a representative production-grade AutoML system. However, FLAML does not natively support fairness or stability metrics, nor multi-stage pipeline optimization; therefore, we extend it with these capabilities for a fair comparison with \sys.
\end{itemize}

\emph{Evaluation of FLAML.}
To ensure a rigorous comparative analysis, it is essential that all systems are evaluated on identical physical hardware. We utilize the \textit{Hyperparameter Tuning (HPT)} module in \flaml, as it is the only component in the framework that natively supports multi-objective optimization. While \flaml offers distributed execution through various backends, we found that \flaml (version 2.3.6) with the Spark backend returns persistent errors when used in conjunction with the HPT module.

Consequently, our benchmarks employ \flaml with the Ray backend. Because multi-node execution for this specific Ray-based configuration is natively optimized for and restricted to proprietary cloud environments (AzureML), conducting a multi-node comparison against our local HPC cluster would introduce significant confounding variables. These include differences in network interconnect latency, disk I/O throughput, and CPU architectures. Therefore, we limit \flaml to single-node execution to ensure that the reported scalability metrics accurately reflect the efficiency of the software architectures rather than variations in the underlying infrastructure.

\subsection{Computing Infrastructure}
\label{apdx:infrastructure}

All experiments were conducted using a suitable experimental environment comprising a high-performance computing (HPC) cluster for execution and an Atlas M10 MongoDB replica set (3 data bearing servers with 2 vCPUs, 2 GB RAM, 1000 IOPS, up to 5 Gigabit network performance) for experiment management. We used the SLURM job scheduler to flexibly assign CPUs, RAM, and nodes for each job on the cluster. All computations were handled by a \texttt{Intel Xeon Platinum 8268 24C 205W 2.9GHz} processor and a \texttt{DDR4 2933MHz} RAM card. Each node provides at most 512 GB of RAM, which is our hardware limit.

Unless stated otherwise, experiments with \sys using 32 workers were run on a single node with fixed resource allocations per dataset below. The same allocations were used for \askl, \alpine, and \flaml.

\begin{itemize}
    \item \diabetes ($\approx$1K): 32 CPUs, 32 GB RAM
    \item \folkemp (15K): 32 CPUs, 64 GB RAM
    \item \folkpc (50K): 32 CPUs, 96 GB RAM
    \item \heart (70K): 32 CPUs, 120 GB RAM
    \item \folkempbig (200K): 32 CPUs, 150 GB RAM
    \item \folkempall ($\approx$2.6M): 32 CPUs, 512 GB RAM
\end{itemize}

The scalability experiment (RQ2.2) in Section~\ref{sec:comparison} allocates one
CPU core per worker and varies RAM with the worker count, as listed below.
All systems received the same allocation at a given worker count. RAM for
\folkempall was tuned to the dataset size and our per-node hardware limit.
For the smaller datasets, allocations are conservative and may exceed what is
strictly required. In this experiment, we report results over 10 random seeds, except for \folkempall, where the cost of a single run limits us to 3 seeds.

\begin{itemize}
    \item \heart (70K): 1, 2, and 4 workers -- 40 GB; 8 and 16 workers -- 60 GB; 32, 64, and 128 workers -- 80 GB per node.
    \item \folkempbig (200K): 1, 2, 4, and 8 workers -- 60 GB; 16 workers -- 80 GB; 32, 64, and 128 workers -- 150 GB per node.
    \item \folkempall ($\approx$2.6M): 4 workers -- 150 GB; 8 workers -- 200 GB; 16 workers -- 256 GB; 32, 64, and 128 workers -- 512 GB per node.
\end{itemize}

All systems were implemented in isolated virtual environments using Python 3.9 and their respective libraries (listed in our repository), based on their original source code and run on Ubuntu 22.04. The code for \alpine was kindly provided by its authors, to whom we express our sincere gratitude. To launch Apache Kafka, we use Singularity containers\footnote{\url{https://docs.sylabs.io/guides/3.5/user-guide/introduction.html}}, deployed on the same node before running an experiment. Specifically, we use the following Docker images: zookeeper --- \texttt{docker://bitnami/zookeeper:3.9.3}, kafka brokers --- \texttt{docker://bitnami/kafka:4.0.0}. Each experiment is repeated ten times to reduce the effect of randomness. All dependencies, hyperparameters, and system configurations are specified in our repository, along with installation and execution instructions in the README.

\section{Additional Experimental Results}
\label{apdx:exp_additional}

\subsection{Multi-Objective Functionality}
\label{apdx:exp1_additional}

Figures~\ref{fig:exp1_folk_pubcov} presents the multi-objective tuning results for \folkpc discussed in Section~\ref{sec:exp1}, complementing the \folkemp and \diabetes results in Figures~\ref{fig:exp1_folk_emp} and \ref{fig:exp1_diabetes}. Points are optimization criteria (OC); values closer to 0 indicate better fairness.

\begin{figure}[b!]
    \centering
    \includegraphics[width=\linewidth]{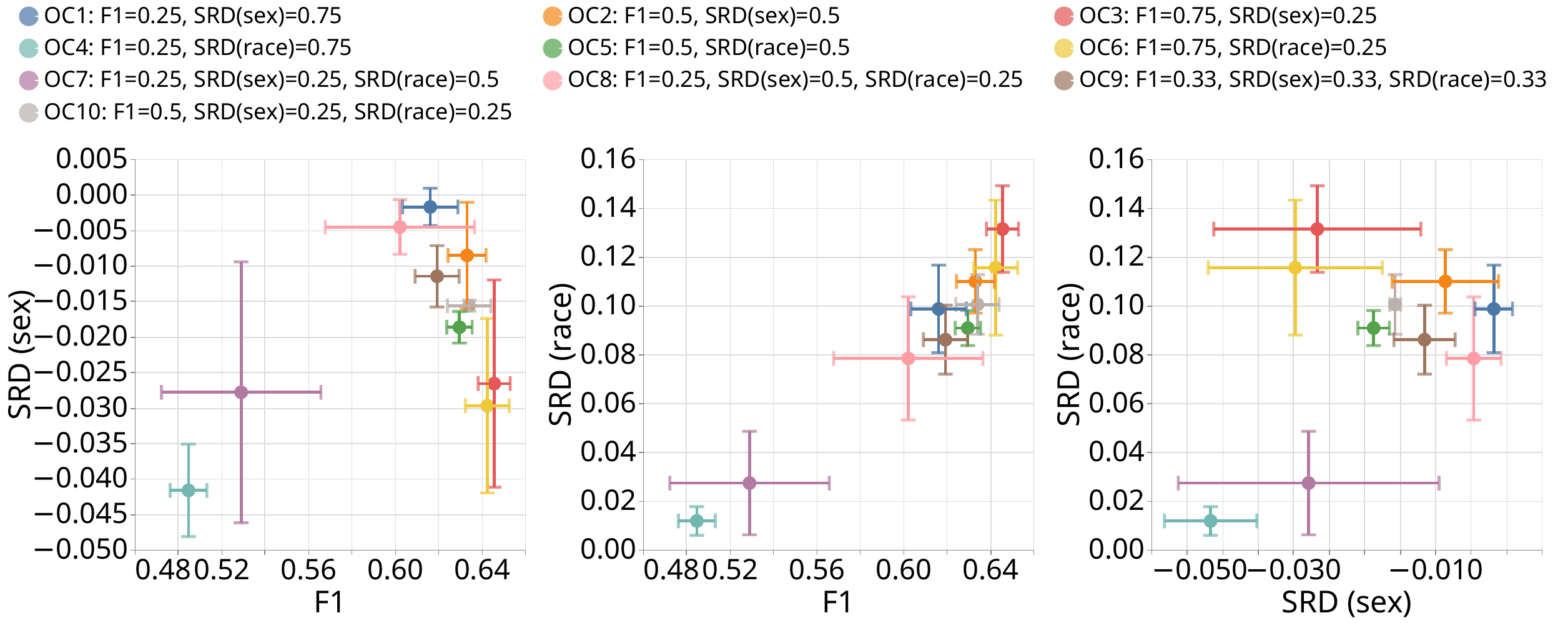}
    \caption{Performance of multi-objective tuning on \folkpc.}
    \label{fig:exp1_folk_pubcov}
\end{figure}


\subsection{Scalability}
\label{apdx:scalability}

Figures~\ref{fig:exp3_all_folk_emp_all}, \ref{fig:exp3_all_heart}, and \ref{fig:exp3_all_folk_emp_big} present supplementary results on model accuracy and fairness for the scalability experiments using the \folkempall, \heart, and \folkempbig datasets, as discussed in Section~\ref{sec:comparison}. These results further support the claim made in the main text that \sys demonstrates clear scalability advantages over existing AutoML systems, particularly in multi-node settings, while maintaining accuracy and fairness as parallelism increases. 

\subsection{System Configuration Sensitivity}
\label{apdx:sensitivity}

\begin{table}[b!]
    \centering
    \footnotesize
    \caption{Sensitivity of \sys to the number of physical pipeline candidates ($k$) per logical pipeline selection.}
    \label{tab:sensitivity_pp}
    \setlength{\tabcolsep}{3pt} 
    \begin{tabular}{rcccc}
        \toprule
        \multirow{2}{*}{\textbf{\# PPs}} & \multicolumn{2}{c}{\textbf{Heart (70K)}} & \multicolumn{2}{c}{\textbf{Folk Pub. Cov.}} \\
        \cmidrule(lr){2-3} \cmidrule(lr){4-5} 
        & Score & Runtime & Score & Runtime \\
        \midrule
        1 & 86.31 \scriptsize{$\pm$0.20} & 1049 \scriptsize{$\pm$461} & 83.05 \scriptsize{$\pm$0.29} & 1411 \scriptsize{$\pm$587} \\
        2 & 86.30 \scriptsize{$\pm$0.21} &  917 \scriptsize{$\pm$180} & 82.90 \scriptsize{$\pm$0.30} & 1095 \scriptsize{$\pm$315} \\
        4 & \cellcolor{gray!20} 86.28 \scriptsize{$\pm$0.33} &  \cellcolor{gray!20} 781 \scriptsize{$\pm$104} & \cellcolor{gray!20} 82.97 \scriptsize{$\pm$0.34} & \cellcolor{gray!20} 1036 \scriptsize{$\pm$284} \\
        8 & 86.17 \scriptsize{$\pm$0.14} &  786 \scriptsize{$\pm$276} & 83.02 \scriptsize{$\pm$0.22} & 1261 \scriptsize{$\pm$430} \\
        16 & 86.08 \scriptsize{$\pm$0.61} &  724 \scriptsize{$\pm$175} & 82.82 \scriptsize{$\pm$0.35} &  909 \scriptsize{$\pm$146} \\
        32 & 85.95 \scriptsize{$\pm$0.91} &  585 \scriptsize{$\pm$165} & 82.91 \scriptsize{$\pm$0.48} &  868 \scriptsize{$\pm$409} \\
        \bottomrule
    \end{tabular}
\end{table}

Table~\ref{tab:sensitivity_halting} presents results for different training set fractions used in pipeline pruning, and Table~\ref{tab:sensitivity_pp} presents results for varying the number of physical pipeline candidates ($k$) per logical pipeline selection, both discussed in Section~\ref{sec:exp4}.

\section{Additional User Study Results}
\label{apdx:user_study_additional}

Table~\ref{tab:user_study_results} provides a detailed breakdown of the user study results discussed in Section~\ref{sec:user_study}, showing correctness and feedback scores per research question aggregated across ten participants. The ``Google Form'' column maps the data to the evaluation forms provided in our accompanying repository\footnote{\url{https://github.com/DataResponsibly/virnyflow-user-study/}}: ``Tasks'' corresponds to {\small\texttt{VirnyFlow\_User\_Study\_Form.pdf}} and ``Feedback'' corresponds to {\small\texttt{VirnyFlow\_User\_Experience\_and\_Usability\_Form.pdf}}. The ``Question'' column indicates the specific section and question number within these documents.

\section{Visualization Interface}
\label{apdx:viz_interfaces}

To enhance user interactivity, \sys includes a dedicated visualization interface designed to support experimentation. It displays the current progress of experiment execution and allows users to explore and interact with historical experimental results. The \interface consists of four pages:

\begin{itemize}[leftmargin=*]
    \item \textbf{Page 1: ``Execution Progress''} (see Figures~\ref{fig:page1_1} and \ref{fig:page1_2}). This page displays a progress bar for the experiment execution, along with score model statistics for each logical pipeline and its associated \textit{experiment config}. It helps users monitor execution status, identify the current best-performing pipeline, and understand the mean and variance of pipeline performance on the task at hand.
    
    \item \textbf{Page 2: ``Pipeline Performance''} (see Figures~\ref{fig:page2_1} and \ref{fig:page2_2}). This page presents a nutritional label~\cite{debull21} for tuned logical pipelines and includes the demographic composition of the dataset. It allows users to assess the performance of the current best pipeline across various dimensions and verify whether the experiment is running correctly.
    
    \item \textbf{Page 3: ``Pipeline Optimization''} (see Figures~\ref{fig:page3_1}-\ref{fig:page3_3}). This page visualizes the optimization progress for each objective. It includes a parallel coordinates plot showing all pipeline executions and their respective objective outcomes, as well as plots for the Pareto frontier, hypervolume, and parameter importance. It helps users explore trade-offs in the optimization process and identify opportunities for improvement.
    
    \item \textbf{Page 4: ``Pipeline Comparison''} (see Figure~\ref{fig:page4_1}). This page enables comparison across logical pipelines and optimizers from different experiments. It also compares system configurations from various \textit{experiment configs}, including their performance scores and runtimes. It helps users select the best-performing optimizer and the most efficient system configuration for a given dataset.
\end{itemize}

\begin{figure}[t!]
    \centering
    \begin{subfigure}{\linewidth}
        \centering
        \includegraphics[width=\linewidth]{AnonymousSubmission/LaTeX/figures/exp3/exp3_speedup_folk_emp_all.pdf}
        \caption{Speedup}
        \vspace{0.25cm}
        \label{fig:exp3_speedup_folk_emp_all}
    \end{subfigure}
    
    \begin{subfigure}{\linewidth}
        \centering
        \includegraphics[width=\linewidth]{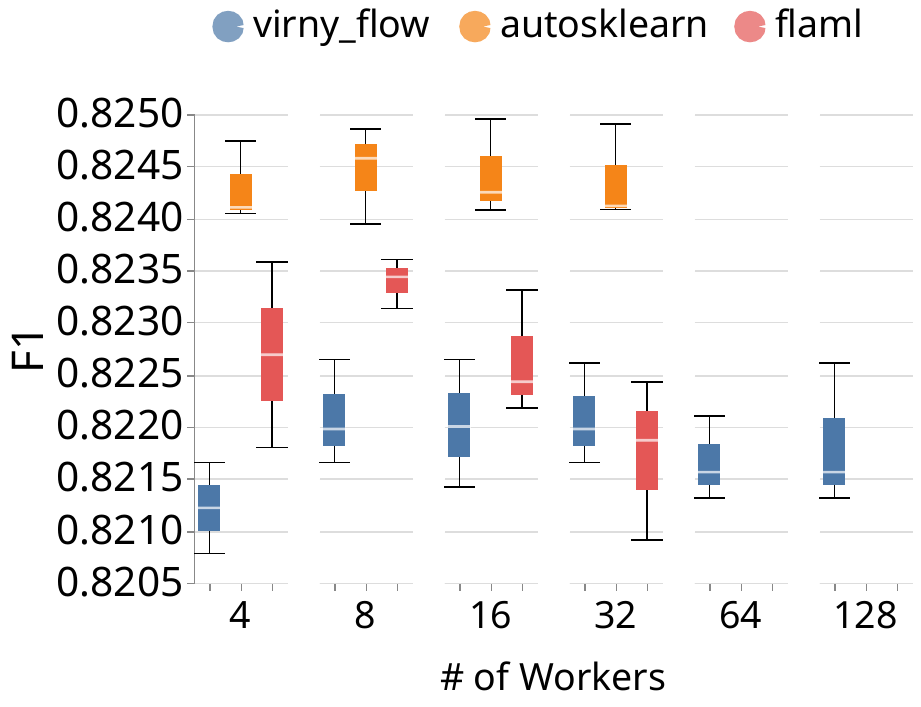}
        \caption{F1}
        \vspace{0.25cm}
        \label{fig:exp3_f1_folk_emp_all}
    \end{subfigure}

    \begin{subfigure}{\linewidth}
        \centering
        \includegraphics[width=\linewidth]{AnonymousSubmission/LaTeX/figures/exp3/exp3_fnrd_folk_emp_all.pdf}
        \caption{FNRD}
        \label{fig:exp3_fnrd_folk_emp_all}
    \end{subfigure}
    
    \caption{Scalability study on \folkempall ($\approx$2.6M). A dashed line highlights the best value for FNRD.}
    \label{fig:exp3_all_folk_emp_all}
\end{figure}

\begin{figure}[t!]
    \centering
    \begin{subfigure}{\linewidth}
        \centering
        \includegraphics[width=\linewidth]{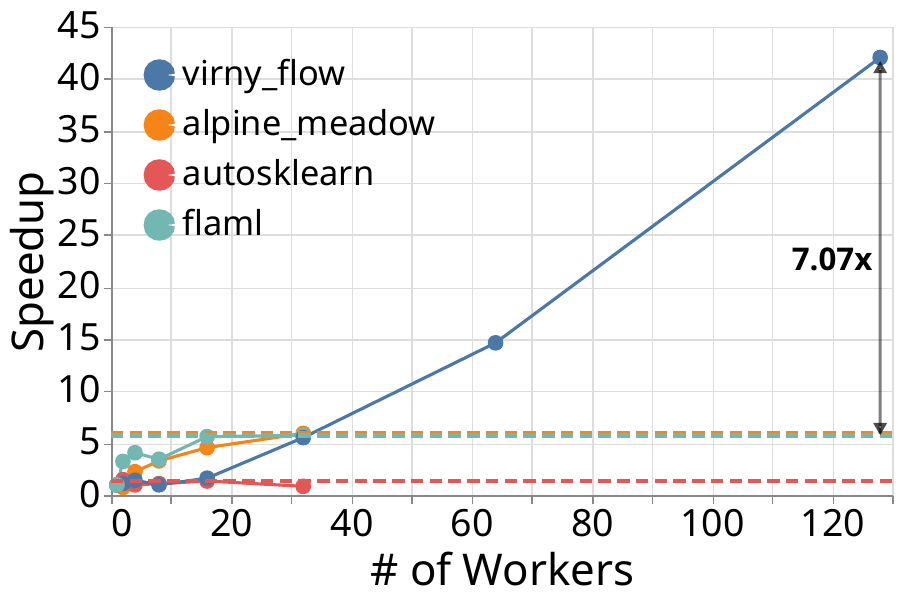}
        \caption{Speedup}
        \vspace{0.25cm}
        \label{fig:exp3_speedup_heart}
    \end{subfigure}
    
    \begin{subfigure}{\linewidth}
        \centering
        \includegraphics[width=\linewidth]{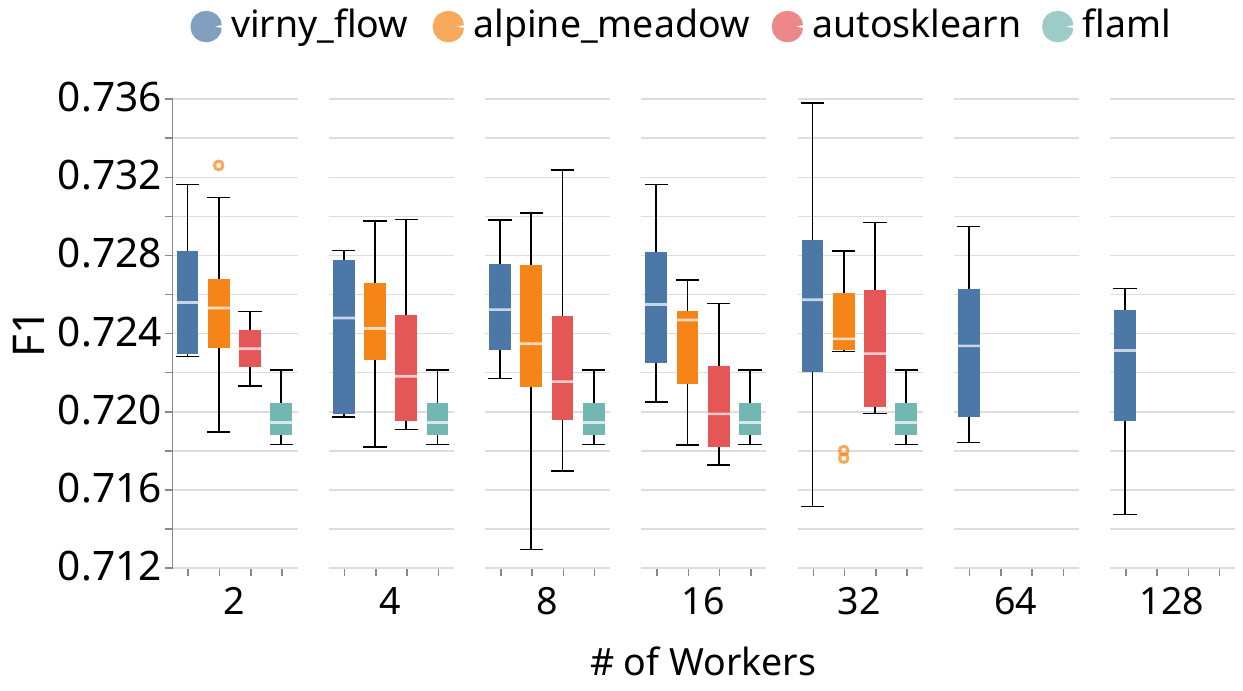}
        \caption{F1}
        \vspace{0.25cm}
        \label{fig:exp3_f1_heart}
    \end{subfigure}

    \begin{subfigure}{\linewidth}
        \centering
        \includegraphics[width=\linewidth]{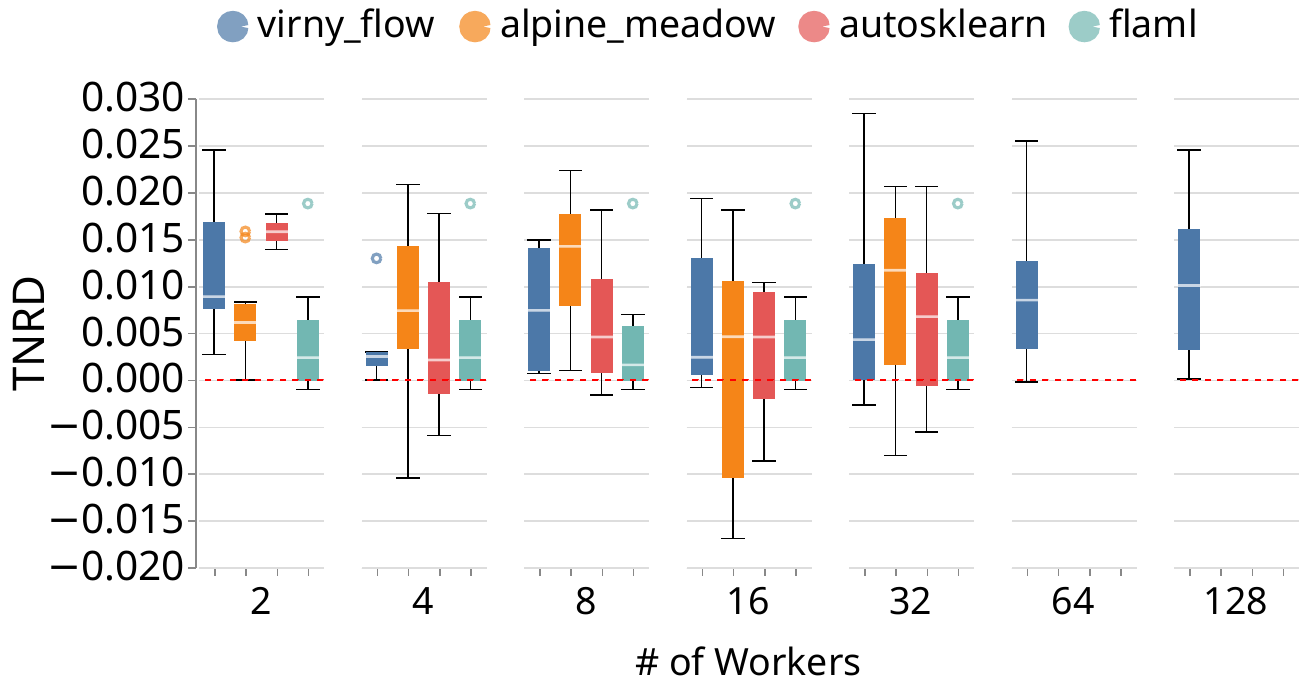}
        \caption{TNRD}
        \label{fig:exp3_tnrd_heart}
    \end{subfigure}
    
    \caption{Scalability study on \heart (70K). A dashed line highlights the best value for TNRD.}
    \label{fig:exp3_all_heart}
\end{figure}

\begin{figure}[t!]
    \centering
    \begin{subfigure}{\linewidth}
        \centering
        \includegraphics[width=\linewidth]{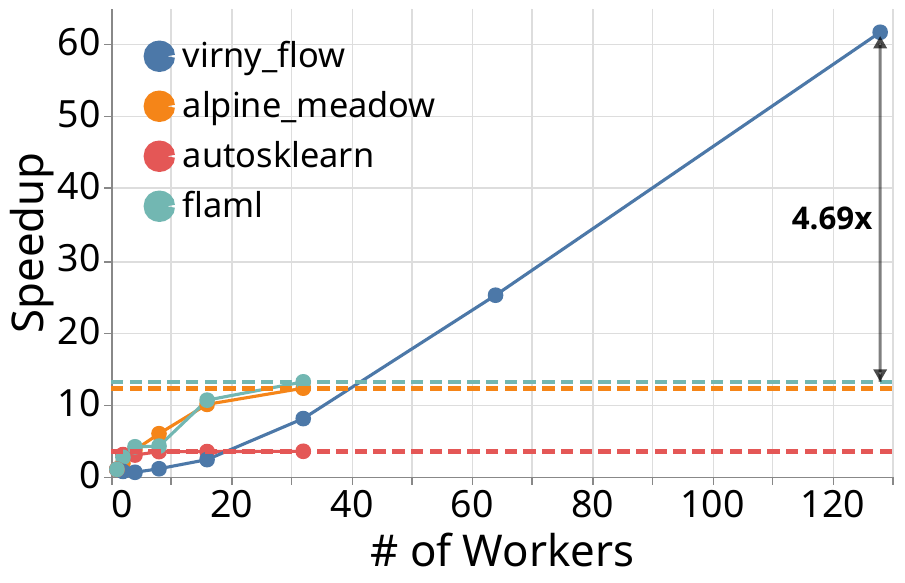}
        \caption{Speedup}
        \vspace{0.25cm}
        \label{fig:exp3_speedup_folk_emp_big}
    \end{subfigure}
    
    \begin{subfigure}{\linewidth}
        \centering
        \includegraphics[width=\linewidth]{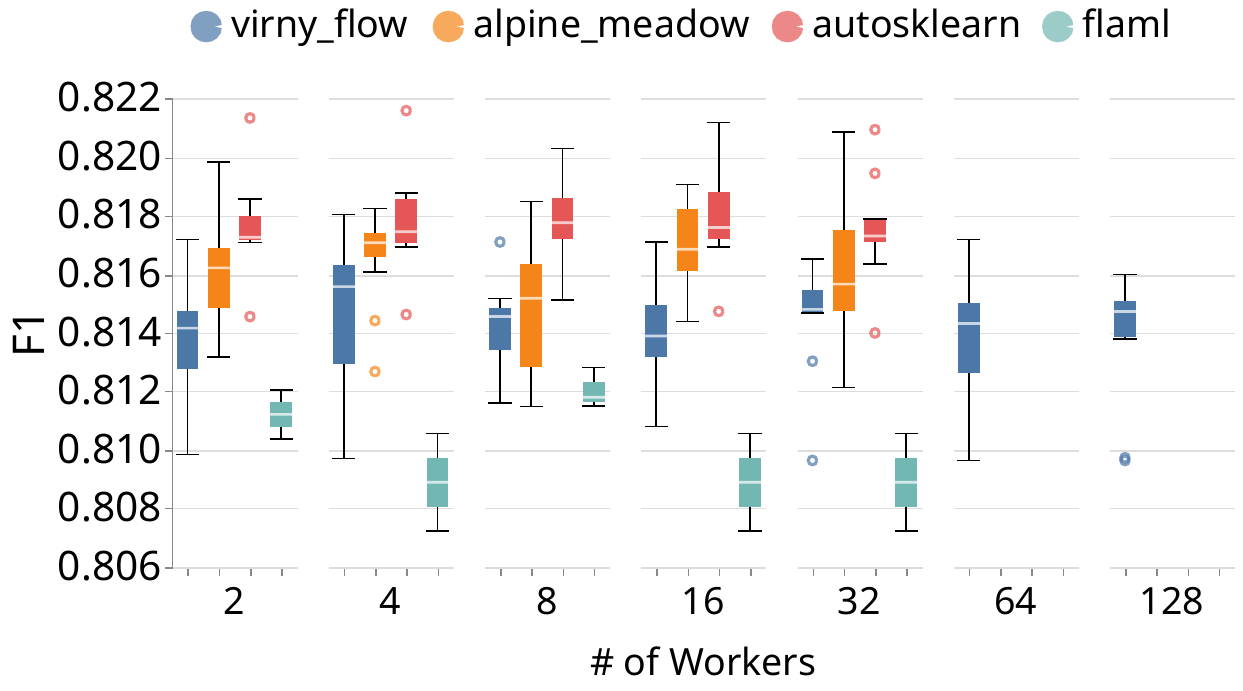}
        \caption{F1}
        \vspace{0.25cm}
        \label{fig:exp3_f1_folk_emp_big}
    \end{subfigure}

    \begin{subfigure}{\linewidth}
        \centering
        \includegraphics[width=\linewidth]{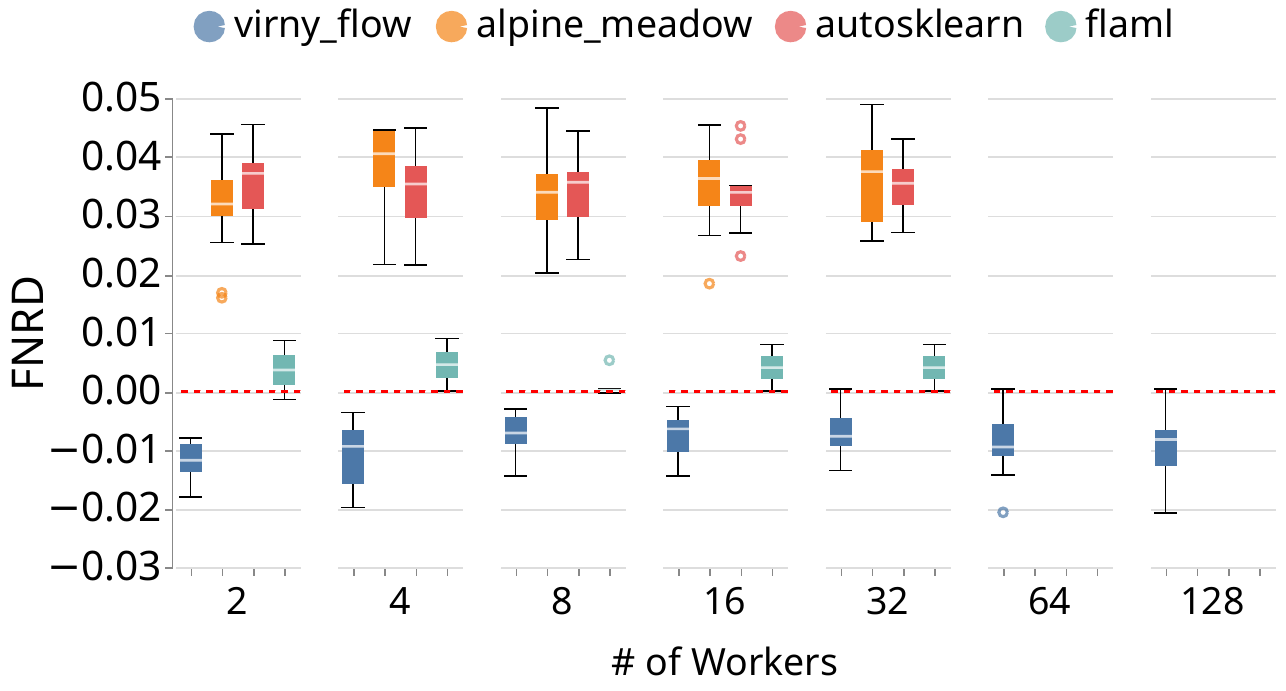}
        \caption{FNRD}
        \label{fig:exp3_fnrd_folk_emp_big}
    \end{subfigure}
    
    \caption{Scalability study on \folkempbig (200K). A dashed line highlights the best value for FNRD.}
    \label{fig:exp3_all_folk_emp_big}
\end{figure}

\begin{table*}[h!]
\centering
\caption{User study results: Correctness and feedback scores per research question aggregated over ten participants.}
\label{tab:user_study_results}
\begin{tabular}{llllr}
\toprule
\textbf{Research Question} & \textbf{Type} & \textbf{Google Form} & \textbf{Question} & \textbf{Result} \\ \midrule

\multirow{8}{*}{Context sensitivity} & \multirow{8}{*}{Quantitative} & \multirow{8}{*}{Tasks} & Section 1: Question 1 & 100\% \\
 &  &  & Section 1: Question 2 & 90\% \\
 &  &  & Section 1: Question 3 & 100\% \\
 &  &  & Section 1: Question 4 & 100\% \\
 &  &  & Section 1: Question 5 & 90\% \\
 &  &  & Section 1: Question 6 & 65\% \\
 &  &  & Section 1: Question 7 & 80\% \\ \cmidrule{4-5} 
 &  &  & \textbf{Total} & \textbf{89.3\%} \\ \midrule

\multirow{12}{*}{Human centricity} & \multirow{12}{*}{Quantitative} & \multirow{12}{*}{Tasks} & Section 2: Question 1 & 93\% \\
 &  &  & Section 2: Question 2 & 100\% \\
 &  &  & Section 2: Question 3 & 100\% \\
 &  &  & Section 2: Question 4 & 100\% \\
 &  &  & Section 2: Question 5 & 90\% \\
 &  &  & Section 2: Question 6 & 100\% \\
 &  &  & Section 2: Question 7 & 100\% \\
 &  &  & Section 2: Question 8 & 100\% \\
 &  &  & Section 2: Question 9 & 90\% \\
 &  &  & Section 2: Question 10 & 100\% \\
 &  &  & Section 2: Question 11 & 85\% \\ \cmidrule{4-5} 
 &  &  & \textbf{Total} & \textbf{96.2\%} \\ \midrule

\multirow{4}{*}{Context sensitivity} & \multirow{4}{*}{Likert} & \multirow{4}{*}{Feedback} & Section 1: Question 2 & 4.2 \\
 &  &  & Section 1: Question 3 & 4.4 \\
 &  &  & Section 2: Question 1 & 4.6 \\ \cmidrule{4-5} 
 &  &  & \textbf{Total} & \textbf{4.4} \\ \midrule

\multirow{5}{*}{Human centricity} & \multirow{5}{*}{Likert} & \multirow{5}{*}{Feedback} & Section 1: Question 4 & 4.1 \\
 &  &  & Section 1: Question 5 & 4.3 \\
 &  &  & Section 2: Question 2 & 4.6 \\
 &  &  & Section 2: Question 3 & 4.1 \\ \cmidrule{4-5}
 &  &  & \textbf{Total} & \textbf{4.275} \\ \midrule

\multirow{4}{*}{System usability} & \multirow{4}{*}{Likert} & \multirow{4}{*}{Feedback} & Section 1: Question 1 & 4.4 \\
 &  &  & Section 3: Question 1 & 4.7 \\
 &  &  & Section 3: Question 2 & 4.7 \\ \cmidrule{4-5} 
 &  &  & \textbf{Total} & \textbf{4.6} \\
\bottomrule
\end{tabular}
\end{table*}

\begin{figure*}[h!]
    \centering
    \includegraphics[width=0.95\linewidth]{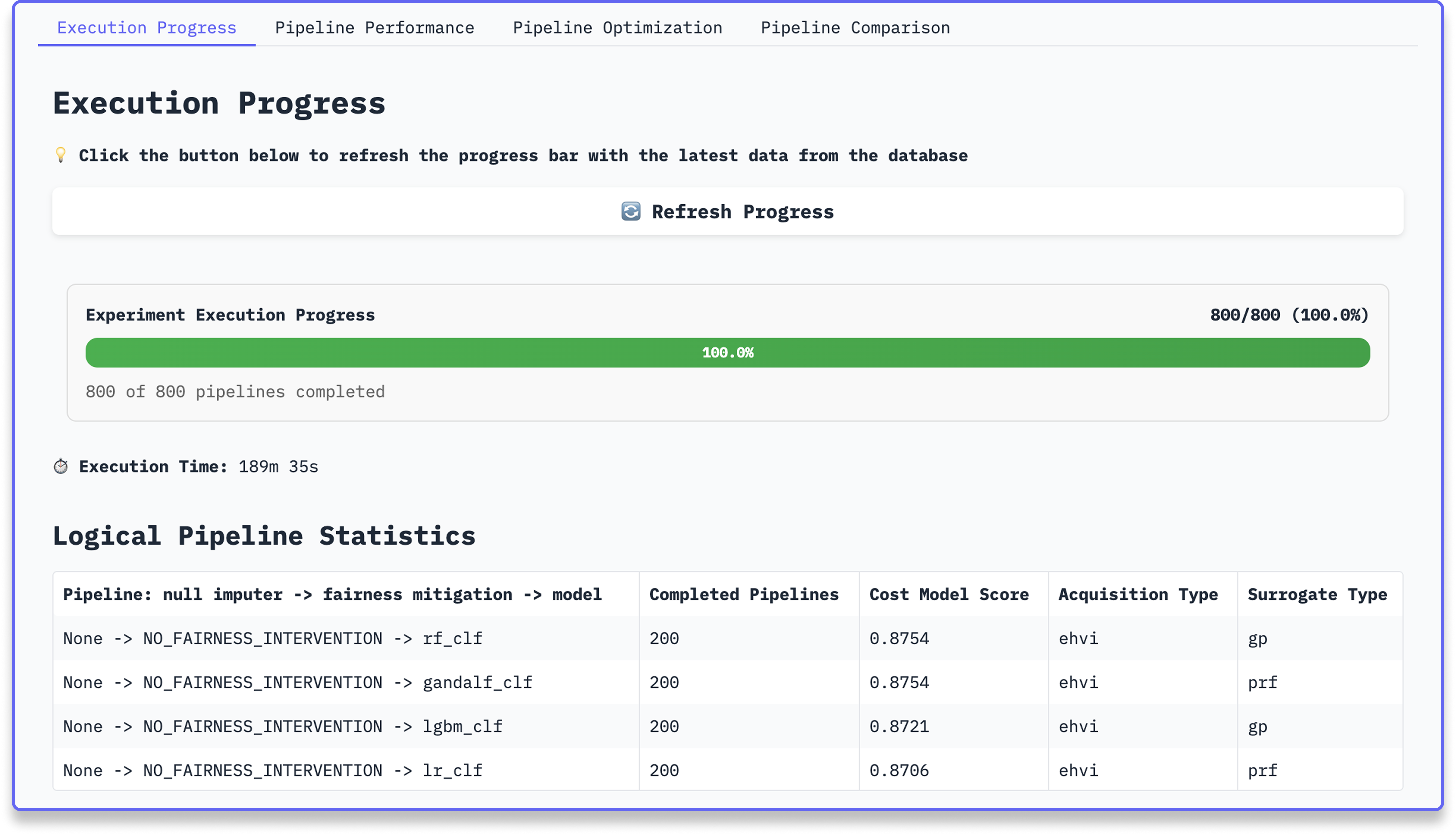}
    \caption{Execution progress page: A progress bar for \folkemp. F1 and Statistical Parity Difference are optimization objectives.}
    \vspace{0.5cm}
    \label{fig:page1_1}
\end{figure*}

\begin{figure*}[h!]
    \centering
    \includegraphics[width=0.85\linewidth]{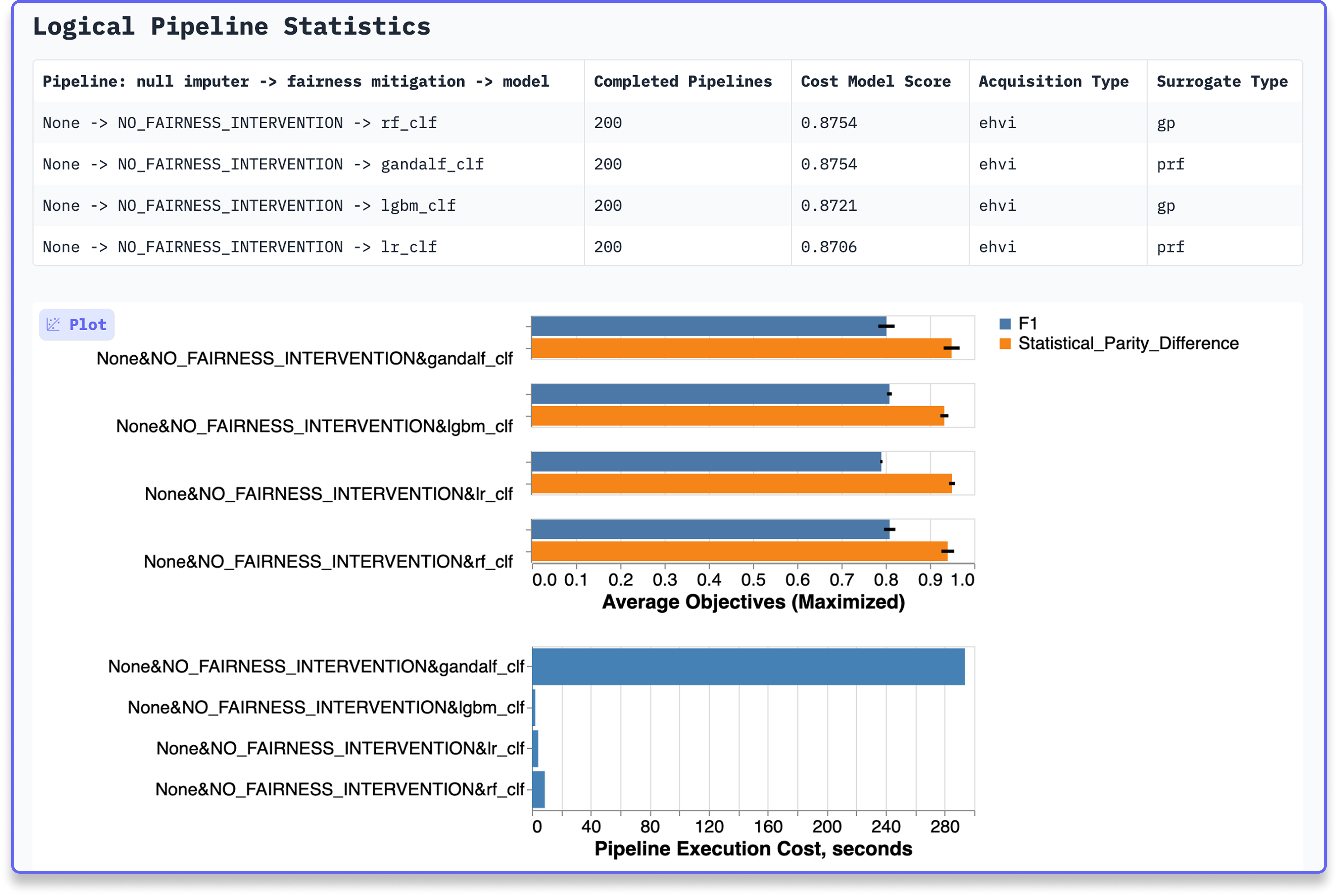}
    \caption{Execution progress page: Logical pipeline statistics for \folkemp.}
    \label{fig:page1_2}
\end{figure*}

\begin{figure*}[h!]
    \centering
    \includegraphics[width=0.75\linewidth]{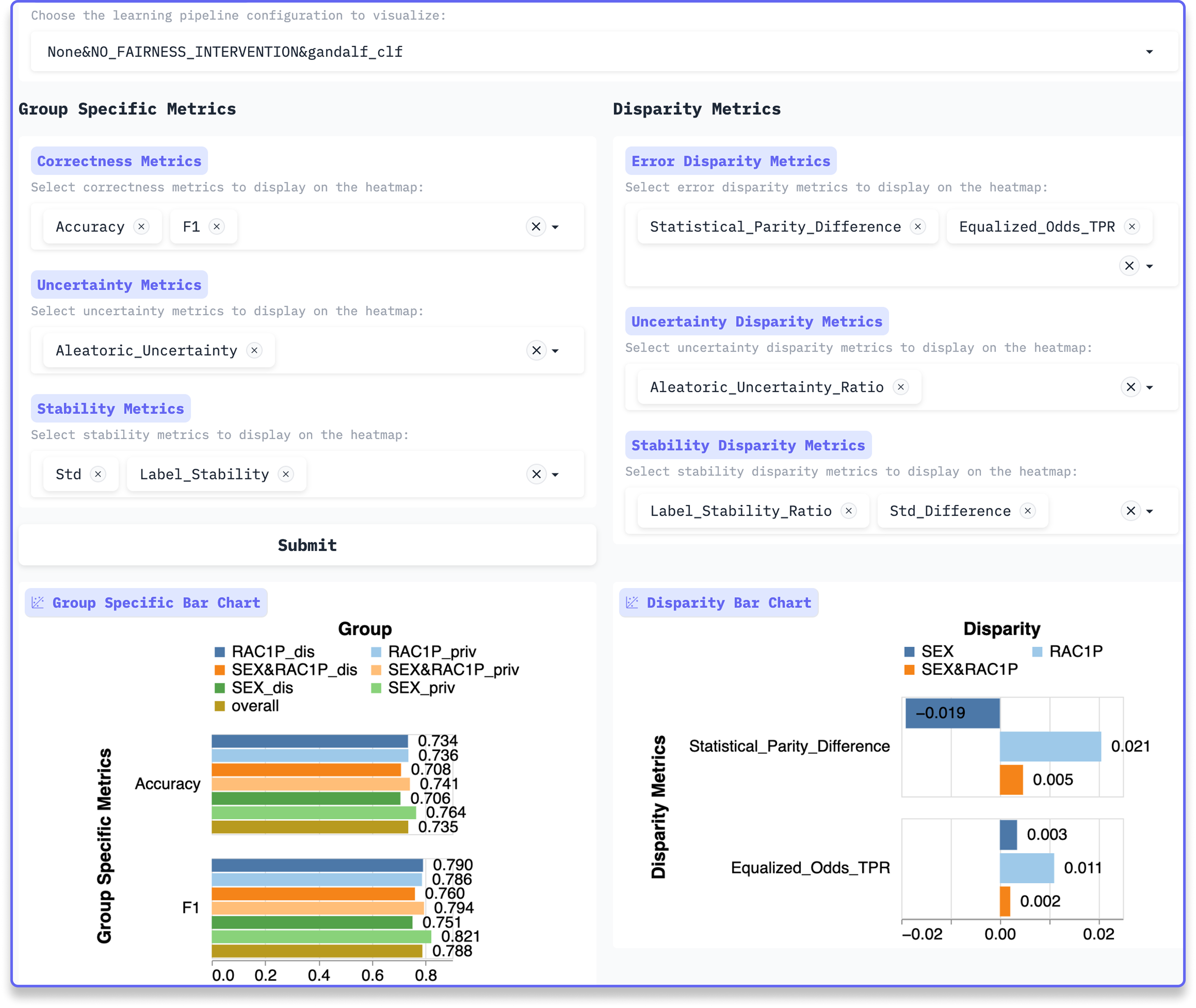}
    \caption{Pipeline performance page: A nutritional label for an ML pipeline with a GANDALF model on \folkemp.}
    \vspace{0.25cm}
    \label{fig:page2_1}
\end{figure*}

\begin{figure*}[h!]
    \centering
    \includegraphics[width=0.75\linewidth]{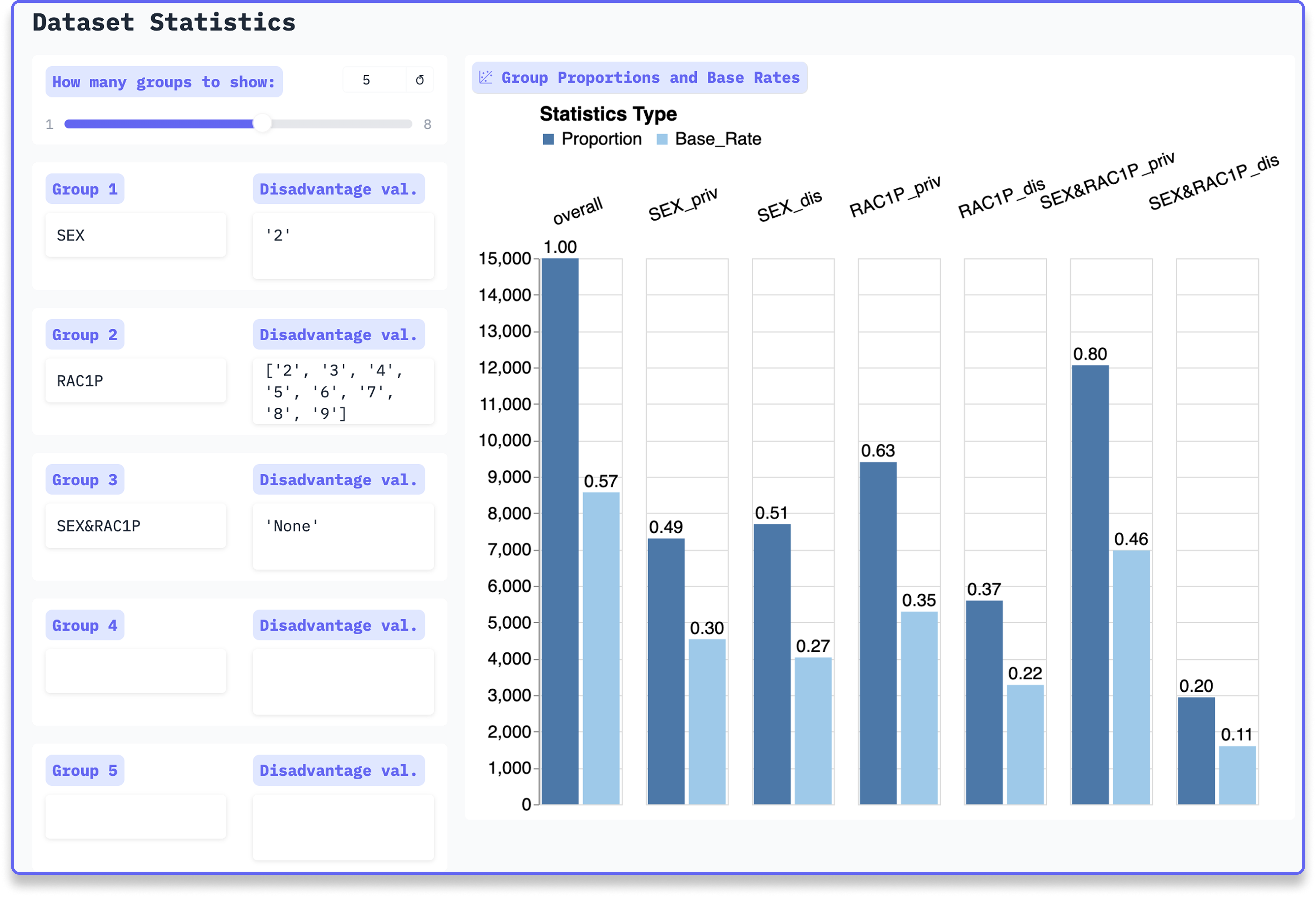}
    \caption{Pipeline performance page: Dataset statistics for \folkemp.}
    \label{fig:page2_2}
\end{figure*}

\begin{figure*}[h!]
    \centering
    \includegraphics[width=0.85\linewidth]{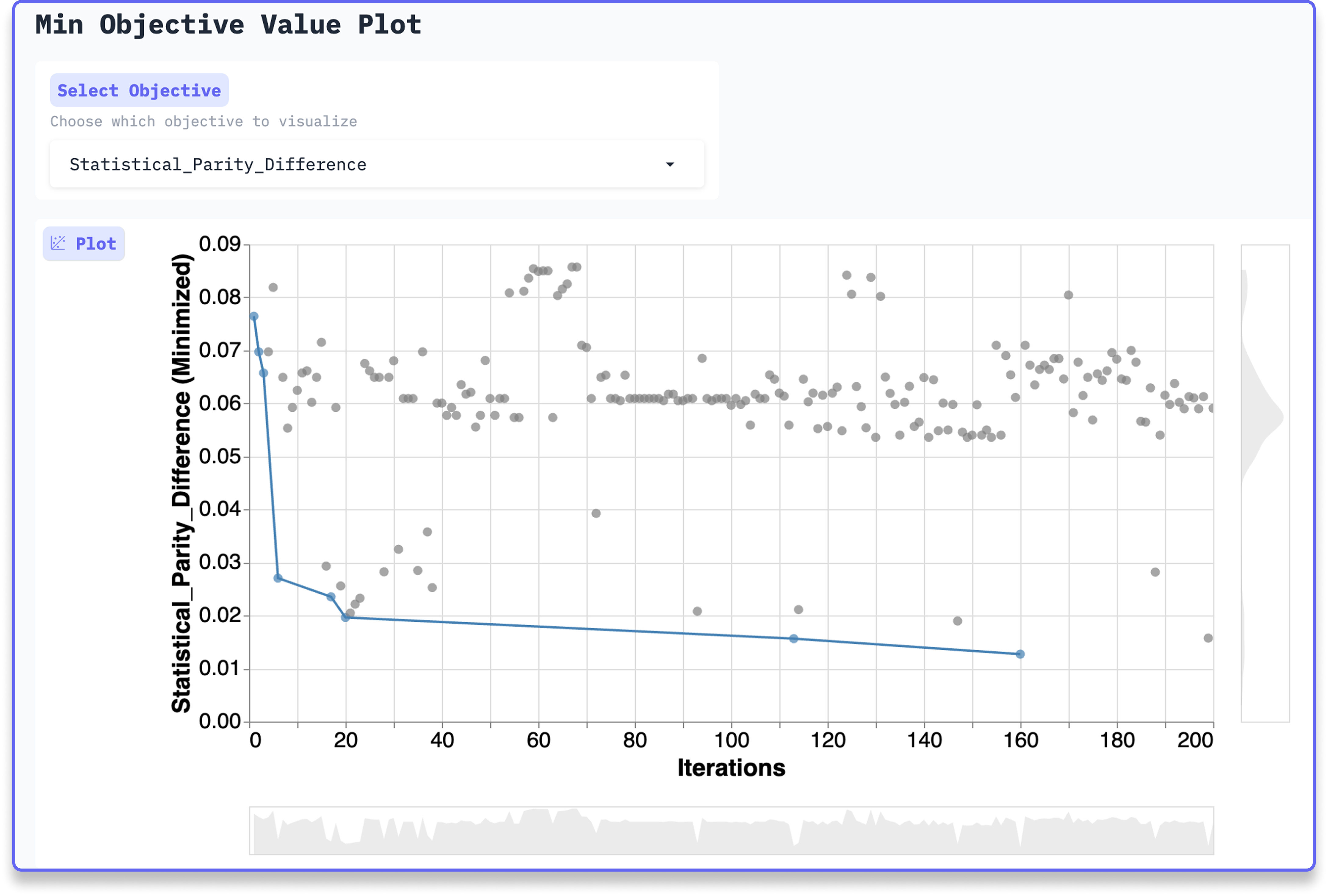}
    \caption{Pipeline optimization page: Optimization progress for Objective 2 (Statistical Parity Difference) for an ML pipeline with Random Forest on \folkemp. The x-axis shows the number of iterations. Since the MOBO optimizer in \sys minimizes the objective, all metrics are transformed accordingly. Lower values indicate better performance.}
    \vspace{0.25cm}
    \label{fig:page3_1}
\end{figure*}

\begin{figure*}[h!]
    \centering
    \includegraphics[width=0.95\linewidth]{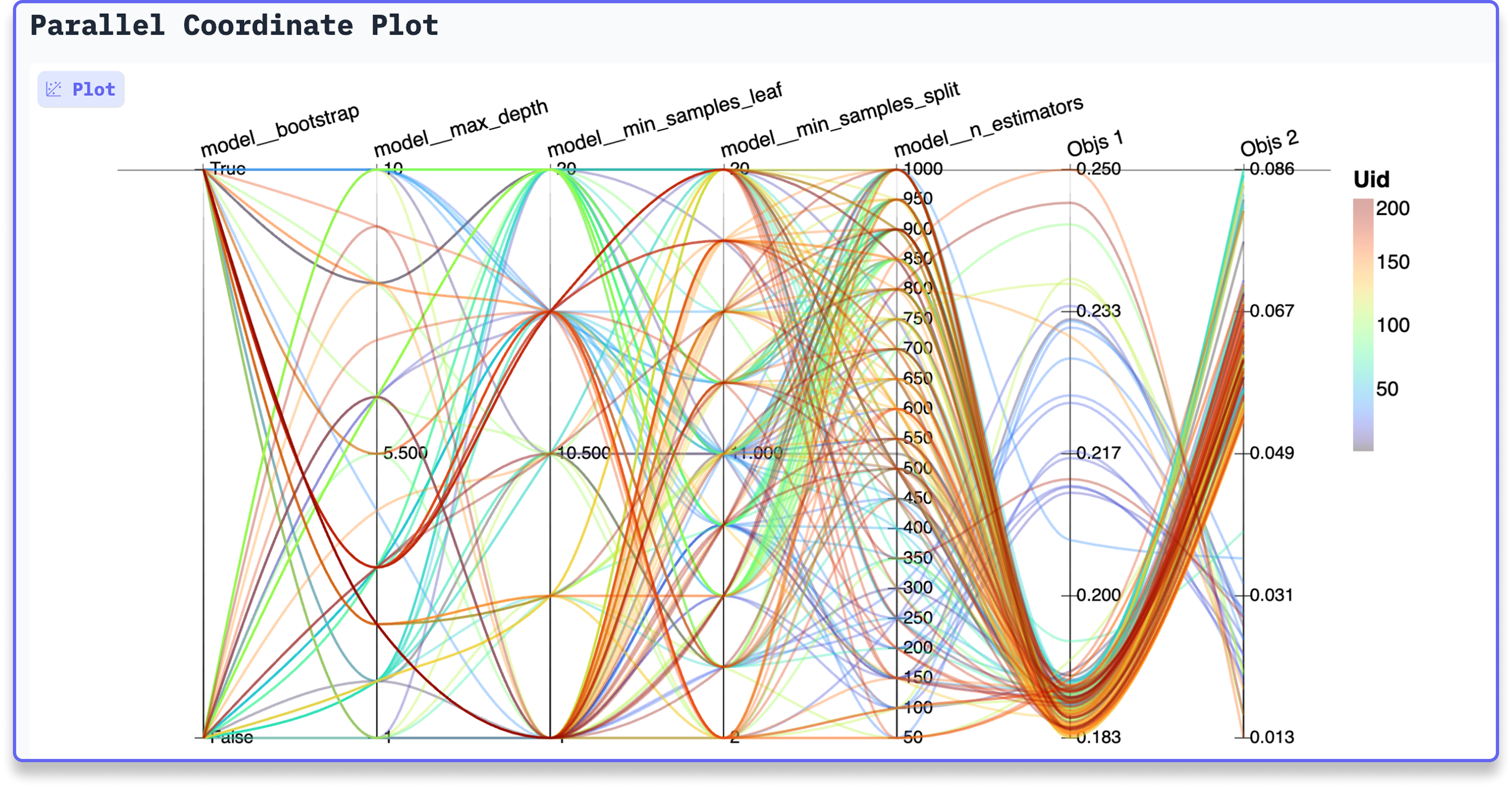}
    \caption{Pipeline optimization page: A parallel coordinates plot for an ML pipeline with Random Forest on \folkemp. The plot shows used hyperparameters for the \lp in each iteration and the respective outcome values of Objectives 1 and 2.}
    \label{fig:page3_2}
\end{figure*}

\begin{figure*}[h!]
    \centering
    \includegraphics[width=0.9\linewidth]{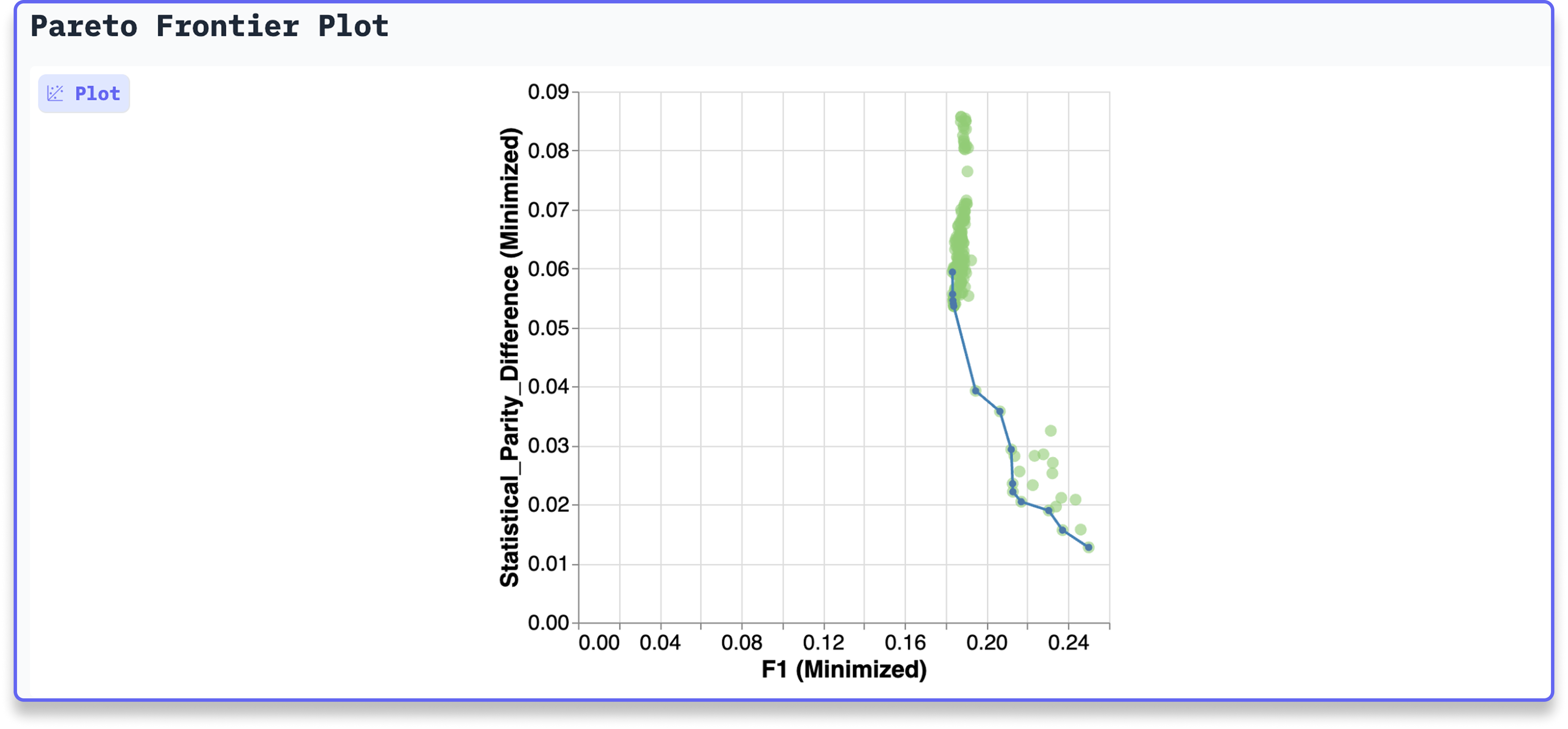}
    \caption{Pipeline comparison page: A Pareto frontier plot for an ML pipeline with Random Forest on \folkemp.}
    \vspace{0.25cm}
    \label{fig:page3_3}
\end{figure*}

\begin{figure*}[h!]
    \centering
    \includegraphics[width=0.85\linewidth]{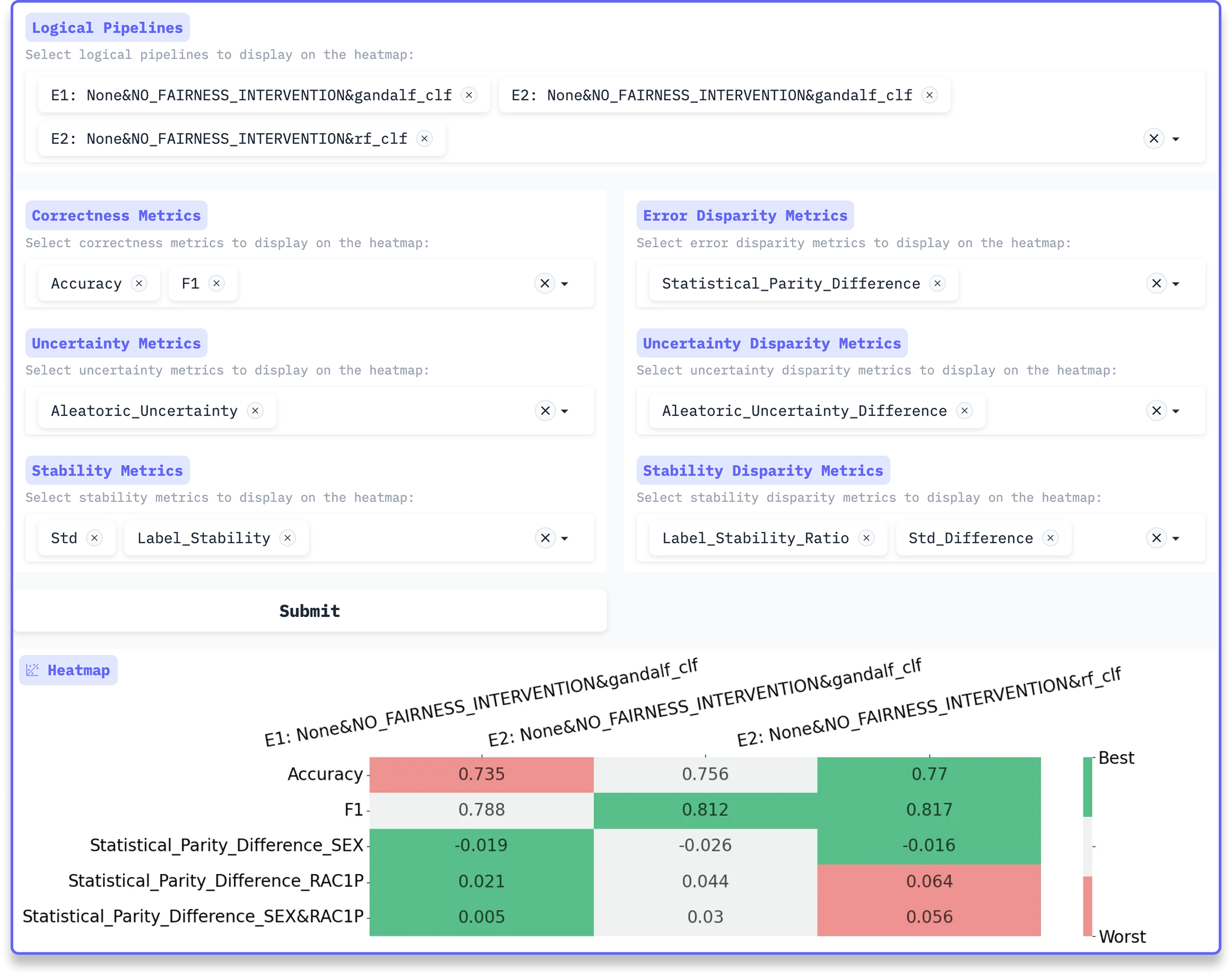}
    \caption{Pipeline comparison page: A comparison heatmap for two different experiment configurations on \folkemp.}
    \label{fig:page4_1}
\end{figure*}

\fi

\end{document}